\documentclass[preprint,12pt]{elsarticle}

\usepackage{amssymb}
\usepackage{amsmath}

\usepackage{algorithm}
\usepackage{algorithmicx}
\usepackage{algpseudocode}
\usepackage{amsfonts}
\usepackage{booktabs}
\usepackage{color}	
\usepackage[table]{xcolor}
\usepackage[a4paper]{geometry}
\usepackage{graphicx}
\usepackage{hyperref}
\usepackage{multirow}
\hypersetup{
	colorlinks=true,
	linkcolor=blue,
	filecolor=magenta,      
	urlcolor=cyan,
	pdftitle={Overleaf Example},
	pdfpagemode=FullScreen,
}

\usepackage{nomencl}
\usepackage[range-phrase=\textup{--}]{siunitx}
\usepackage{subcaption}
\usepackage{svg}
\usepackage{tabularray}
\usepackage{wrapfig}

\DeclareMathOperator*{\argmin}{\mathrm{argmin}}
\DeclareMathOperator*{\argmax}{\mathrm{argmax}}

\journal{Applied Mathematical Modelling}

\begin{document}
	\sloppy
	
	\begin{frontmatter}
		
		
		
		\title{Adaptive Bayesian Data-Driven Design of Reliable Solder Joints for Micro-electronic Devices}
		
		
		\author[Delft-EEMCS]{Leo Guo} 
		
		\author[Delft-EEMCS]{Adwait Inamdar} 
		
		\author[Delft-EEMCS]{Willem D. van Driel} 
		
		\author[Delft-EEMCS]{GuoQi Zhang\corref{cor1}} 
		\ead{g.q.zhang@tudelft.nl}
        
		\cortext[cor1]{Corresponding author.}
		
		
		\affiliation[Delft-EEMCS]{organization={Department of Electronic Components, Technology and Materials, Delft University of Technology},
			addressline={Mekelweg 4}, 
			city={Delft},
			postcode={2628CM},
			country={Netherlands}}
		
		\begin{abstract}
			Solder joint reliability related to failures due to thermomechanical loading is a critically important yet physically complex engineering problem. As a result, simulated behavior is oftentimes computationally expensive. In an increasingly data-driven world, the usage of efficient data-driven design schemes is a popular choice. Among them, Bayesian optimization (BO) with Gaussian process regression is one of the most important representatives. The authors argue that computational savings can be obtained from exploiting thorough surrogate modeling and selecting a design candidate based on multiple acquisition functions. This is feasible due to the relatively low computational cost, compared to the expensive simulation objective. This paper addresses the shortcomings in the adjacent literature by providing and implementing a novel heuristic framework to perform BO with adaptive hyperparameters across the various optimization iterations. Adaptive BO is subsequently compared to regular BO when faced with synthetic objective minimization problems. The results show the efficiency of adaptive BO when compared any worst-performing regular Bayesian schemes. As an engineering use case, the solder joint reliability problem is tackled by minimizing the accumulated non-linear creep strain under a cyclic thermal load. Results show that adaptive BO outperforms regular BO by 3\% on average at any given computational budget threshold, critically saving half of the computational expense budget. This practical result underlines the methodological potential of the adaptive Bayesian data-driven methodology to achieve better results and cut optimization-related expenses. Lastly, in order to promote the reproducibility of the results, the data-driven implementations are made available on an open-source basis.
		\end{abstract}
		
		
	\end{frontmatter}
	
	\section{Introduction}
	
	Computer simulations of integrated circuit packages, particularly finite element model (FEM) simulations have become an effective tool in improving their designs. Extensive use of data-driven design in addition to the traditional physics-based design process has been a central theme in engineering for the past years \cite{guo2021artificial}. The simulations that underpin the design process, however, can be computationally expensive. One way to handle this is by leveraging data to construct related sub-objectives that are much faster and simpler to evaluate and optimize. One of the most popular method that satisfies these properties is Bayesian optimization (BO) with Gaussian process regression (GPR) \cite{shahriari2015taking,garnett2023bayesian}. In short, BO is a proxy-optimization method, which employs knowledge from a surrogate model to do adaptive design space sampling in order to optimize an expensive objective. While many types of surrogate models can be selected, the surrogate is commonly chosen to be a Gaussian process (GP) regression model. The proxy-optimization component of BO hinges on the selection of an acquisition function, which maps the design space onto a belief landscape that serves to guide the adaptive sampling process. As a result, BO with GP regression models has proven to be a fruitful methodology in the efficient design of micro-electronic components \cite{zhang2019efficient,huang2022bayesian,zhang2021efficient}.
	
	The modern societal importance of robust micro-electronics is evident. However, electronic components undergo degradation under the environmental and operating loads, which leads to package-level and board-level failures. These failures occur either due to an event of a sudden change in loads (such as, excess temperature, excess current or voltage, mechanical shock, stress or impact) or a prolonged exposure to nominal operating conditions. About 70\% of the failures in electronic components occur during the packaging and assembly processes, and the predominant failure mode is associated with the solder joints \cite{tilgner2009physics}. Temperature, humidity, mechanical vibrations, and dust are the four key environmental factors that are responsible for component degradation, in which the temperature factor is the most dominant one \cite{li2017review,musadiq2025impact} and is responsible for about 55\% of the failures; whereas mechanical vibrations contribute to about 20\% of the failures \cite{pecht2018handbook}. Failures related to semiconductors, connectors, and solder joints together account for over one-third of the share for power electronics \cite{wolfgang2007examples}. A solder joint failure is primarily governed by the variation of temperature and mechanical loads \cite{ogbomo2018effect, arabi2020effect}, and thus, it is one of the key aspects of reliability engineering for integrated circuit packages and electronics-enabled systems. Chief among the indicators of solder joint failure is accumulated non-linear creep strain \cite{wong2016creep}, which can be calculated by means of FEM simulations. However, numerically modeling the thermomechanical behavior of solder joints is often a computationally expensive endeavor. In order to compensate for this expense, response surface modeling of accumulated (plastic) strain has previously been achieved by means of long short-term memory neural network models \cite{de2024physics,de2025solder}. This work served as a foundation to utilize GPR in modeling non-linear creep \cite{van2003response}, and was previously adapted with BO to solve for reliability in solder joints \cite{wymyslowski2007advanced}.
	
	It is a challenge to tackle design problems in a data-driven manner while keeping high computational expense of the objective in mind. This is exacerbated by the fact that response surfaces in engineering problems can be noisy or multimodal \cite{frazier2018tutorial}. It is therefore important to thoroughly discuss the hyperparameters or model parameters which BO with GPR hinge on. Practitioners of supervised machine learning, such as regression, are familiar with the importance of obtaining a model with a model parameters vector $\boldsymbol{\theta}$, that generalizes the regression model across the entire design space as well as possible \cite{feurer2015initializing,yang2020hyperparameter}. Concretely, this is commonly undertaken by splitting the DoE $\textbf{D}$ into train ($\textbf{D}_{\text{train}}$), validation ($\textbf{D}_{\text{val}}$) and test sets ($\textbf{D}_{\text{test}}$). Subsequently, a GP regression model is constructed over $\textbf{D}_{\text{train}}$. In the case of $k$-fold cross-validation, $k$ different train-validation splits are made, resulting in a set of $k$ model parameter vector candidates $\hat{\boldsymbol{\theta}}_1,\ldots,\hat{\boldsymbol{\theta}}_k$, out of which the best performing model parameter vector is chosen. Lastly, the capability of the resulting regression model to generalize the prediction of $f$ is tested by calculating its performance with $\textbf{D}_{\text{test}}$. This workflow has been successfully used in supervised machine learning assisted engineering applications, such as medical tomography \cite{ueda2021training}, energy consumption footprints in construction \cite{khalil2022machine} and polymer modeling \cite{sharma2022advances}. Crucially, to the authors' knowledge, the handling of supervised model hyperparameters in the context of Bayesian data-driven design lacks representation in the adjacent literature. Finally, it is noted that the acquisition function type is commonly kept the same throughout all steps of BO. This is despite the lack of precedent in believing that other available acquisition function types would perform worse. Due to the high stakes that the high-cost objective evaluations carry over to the optimization problem, it is a critically important issue to address.
	
	
	This work aims to show that it is possible to create statistically robust heuristics towards the selection of hyperparameters when performing BO, to the benefit of efficient usage of a given computational budget. First, a brief introduction of BO with GPR is given, after which the novel ideas of surrogate model selection and acquisition candidate selection are introduced. With regards to the micro-electronic case study, variance-based sensitivity analysis is performed to identify the relevant design parameters that should be considered for an optimization problem. Finally, the adaptive BO framework is applied to optimize a solder joint design for minimal accumulated creep strain and compared to the non-adaptive, standard variant of BO. Because the authors value the accessibility and reproducibility of the results, the code implementations and data-related resources are open-source as mentioned in the Data availability statement.
	
	\section{Bayesian data-driven methods}
	
	\subsection{Bayesian optimization with Gaussian process regression}
	
	GP modelling of an objective function $f:[0,1]^D\to\mathbb{R}$ is a Bayesian method, in that it assumes all dependent variables to be stochastic \cite{williams2006gaussian}. To be precise, for any $\textbf{x}\in[0,1]^D$, the value $f(\textbf{x})$ is modelled as a stochastic variable. For any $\textbf{u},\textbf{v}\in[0,1]^D$, the covariance between $f(\textbf{u})$ and $f(\textbf{v})$ is stipulated by a covariance function or kernel $\kappa$, such that $\kappa(\textbf{u},\textbf{v}):=\text{Cov}(f(\textbf{u}),f(\textbf{v}))$. Various choices for $\kappa$ exist, each of them relying on a vector of learnable model parameters $\boldsymbol{\theta}$. The most popular ones in literature are:
	\begin{align}
		\begin{split}
			\kappa_{\text{RBF},\boldsymbol{\theta}_{\text{RBF}}}(\textbf{u},\textbf{v})&:=c\cdot\exp\left(-\frac{r^2(\textbf{u},\textbf{v})}{2\lambda^2}\right)+s^2\delta(\textbf{u}-\textbf{v}),\\
			\kappa_{\text{Mat},\boldsymbol{\theta}_{\text{Mat}}}(\textbf{u},\textbf{v})&:=c\cdot\left(1+\frac{\sqrt{3}r(\textbf{u},\textbf{v})}{\lambda}\right)\exp\left(-\frac{\sqrt{3}r(\textbf{u},\textbf{v})}{\lambda}\right)+s^2\delta(\textbf{u}-\textbf{v}),\\	\kappa_{\text{RQ},\boldsymbol{\theta}_{\text{RQ}}}(\textbf{u},\textbf{v})&:=c\cdot\left(1+\frac{r^2(\textbf{u},\textbf{v})}{2\alpha\lambda^2}\right)^{-\alpha}+s^2\delta(\textbf{u}-\textbf{v}),
		\end{split}
		\label{eq:kernels}
	\end{align}
	where $r(\textbf{u},\textbf{v}):=\|\textbf{u}-\textbf{v}\|$ and $\delta$ is the Dirac-$\delta$ function. Furthermore, the scalars $c,\lambda,s^2,\alpha$ represent covariance kernel parameters, and are summarized into a model parameter vector per kernel type, generically denoted by $\boldsymbol{\theta}$. For example, the RBF kernel has $\boldsymbol{\theta}_{\text{RBF}}:=(c,\lambda,s^2)^\top$ as its model parameter vector.
	
	Furthermore, assume that $\textbf{X}:=(\textbf{x}_1^\top,\ldots,\textbf{x}_N^\top)^\top$ is a matrix of $N$ design parameter vectors, then $\textbf{y}:=(f(\textbf{x}_1),\ldots,f(\textbf{x}_N))^\top$ is the realization of a multivariate random variable. If one now assumes that $\boldsymbol{\theta}$, as a dependent parameter, is a realization of a random variable $\boldsymbol{\Theta}$, and
	\begin{equation}
		\textbf{Y}|(\boldsymbol{\Theta}=\boldsymbol{\theta})\sim\mathcal{N}(\textbf{0},\textbf{K}_{\boldsymbol{\theta}})
		\label{gp-prior}
	\end{equation}
	with covariance matrix $\textbf{K}_{\boldsymbol{\theta}}=\textbf{K}_{\boldsymbol{\theta}}(\textbf{X}):=(\kappa_{\boldsymbol{\theta}}(\textbf{x}_i,\textbf{x}_j))_{i,j=1,\ldots,N}$, then
	\begin{equation}
		f(\textbf{x})|(\textbf{Y}=\textbf{y},\boldsymbol{\Theta}=\boldsymbol{\theta})\sim\mathcal{N}(\mu_{\boldsymbol{\theta}}(\textbf{x}),\sigma_{\boldsymbol{\theta}}^2(\textbf{x})),
		\label{eq:intermediate-ppd}
	\end{equation}
	where
	\begin{align}
		\mu_{\boldsymbol{\theta}}(\textbf{x})&:=\kappa_{\boldsymbol{\theta}}(\textbf{x},\textbf{X})^\top\textbf{K}_{\boldsymbol{\theta}}^{-1}\textbf{y},
		\label{gp-ppd-mean}\\
		\sigma_{\boldsymbol{\theta}}^2(\textbf{x})&:=\kappa_{\boldsymbol{\theta}}(\textbf{x},\textbf{x})-\kappa_{\boldsymbol{\theta}}(\textbf{x},\textbf{X})^\top \textbf{K}_{\boldsymbol{\theta}}^{-1}\kappa_{\boldsymbol{\theta}}(\textbf{x},\textbf{X}).
		\label{gp-ppd-var}
	\end{align}
	Selecting or finding a fitting value for $\boldsymbol{\theta}$ is called GP regression (GPR). One common method of doing so is by numerically solving for the maximum (log) likelihood estimate (MLE):
	\begin{equation}
		\hat{\boldsymbol{\theta}}_{\text{MLE}}:=\argmin_{\boldsymbol{\theta}\in{\mathcal{T}_\kappa}}\ln(\det(\textbf{K}_{\boldsymbol{\theta}}(\textbf{X})))+\textbf{y}^\top \textbf{K}^{-1}_{\boldsymbol{\theta}}(\textbf{X})\textbf{y}.
		\label{eq:mle}
	\end{equation}
	where $\mathcal{T}_\kappa$ stands for the space of all permissible model parameter vectors $\boldsymbol{\theta}$. The normal distribution that results from inserting $\hat{\boldsymbol{\theta}}_{\text{MLE}}$, i.e. $\mathcal{N}(\mu_{\hat{\boldsymbol{\theta}}_{\text{MLE}}}(\textbf{x}),\sigma_{\hat{\boldsymbol{\theta}}_{\text{MLE}}}^2(\textbf{x}))$, is called the regressive-predictive distribution (RPD). As a normal distribution, an RPD is fully described by its probability density function, which will be denoted by $\hat{\phi}$. Accordingly, the mean and variance of this distribution are denoted as $\hat{\mu}:=\mu_{\hat{\boldsymbol{\theta}}_{\text{MLE}}}$ and $\hat{\sigma}^2:=\sigma^2_{\hat{\boldsymbol{\theta}}_{\text{MLE}}}$.
	
	In the assumption that $f$ is to be numerically minimized, an RPD $\hat{\phi}$ carries with it valuable knowledge to suggest new design parameter vectors to sample $f$ at. A common way to extract this knowledge is to build an acquisition function $\alpha:[0,1]^D\to\mathbb{R}$ such that $\alpha(\textbf{x};\hat{\phi})$ quantifies a level of belief that $f(\textbf{x})$ is less than any component of $\textbf{y}$.
	
	An example of an acquisition function $\alpha$ is the expected improvement (EI) acquisition \cite{jones1998efficient}. 
	Other popular examples that have been utilized in data-driven literature include probability of improvement (PI) \cite{kushner1964new} and lower / upper confidence bound (UCB) \cite{auer2002using} with hyperparameter $\beta$. See Equation \eqref{eq:acqs}.
	\begin{align}
		\begin{split}
			\alpha_{\text{EI}}(\textbf{x};\hat{\phi})&:=\hat{\sigma}(\textbf{x})(z(\textbf{x})\hat{\Phi}(z(\textbf{x}))+\hat{\phi}(z(\textbf{x})));\\
			\alpha_{\text{PI}}(\textbf{x};\hat{\phi})&:=\hat{\Phi}(z(\textbf{x}));\\
			\alpha_{\text{UCB}}(\textbf{x};\hat{\phi},\beta)&:=\hat{\mu}(\textbf{x})+\beta\hat{\sigma}(\textbf{x}).
		\end{split}
		\label{eq:acqs}
	\end{align}
	By numerically optimizing (maximizing) $\alpha$ across the design parameter domain, a promising design parameter vector $\textbf{x}^*$ can be suggested to evaluate $f(\textbf{x}^*)$. It should be noted that the formulations in Equation \eqref{eq:acqs} are all analytical and differentiable, meaning that gradient-based optimizers such as Adam \cite{kingma2014adam} and L-BFGS \cite{zhu1997algorithm} may be employed. In the case of EI and PI, these gradient-based schemes may not always converge quickly, so the enhanced logarithmic EI (LogEI) \cite{ament2023unexpected} and logarithmic PI are often used as practical alternatives.
	
	By appending $\textbf{X}$ with $\textbf{x}^*$ and $\textbf{y}$ with $f(\textbf{x}^*)$, the process of obtaining a renewed surrogate model can start anew. This process as a whole is called BO with GPR. See Algorithm \ref{algo:bo} for an overview. 
	
	\begin{algorithm}[H]
		\caption{Bayesian optimization with Gaussian process regression}
		\begin{algorithmic}[1]
			\Require{Design of training experiments $\textbf{D}^{(0)}=(\textbf{X}^{(0)},\textbf{y}^{(0)})$, covariance function $\kappa$, acquisition function $\alpha$, number of iterations $I$}
			\For{$i=1,\ldots,I$}
			\State\footnotesize$\hat{\boldsymbol{\theta}}_{\text{MLE}}^{(i)}\gets\argmin_{\boldsymbol{\theta}\in\mathcal{T}_\kappa}\ln(\det(\textbf{K}_{\boldsymbol{\theta}}(\textbf{X}^{(i-1)})))+\textbf{y}^{(i-1)\top} \textbf{K}^{-1}_{\boldsymbol{\theta}}(\textbf{X}^{(i-1)}) \textbf{y}^{(i-1)}$\Comment{Eq. \eqref{eq:mle}}
			\normalsize\State $\hat{\phi}^{(i)}\gets\hat{\boldsymbol{\theta}}_{\text{MLE}}^{(i)}$
			\State $\textbf{x}^{(i)}\gets\argmax_{\textbf{x}\in[0,1]^D}\alpha(\textbf{x};\hat{\phi}^{(i)})$
			\State $y^{(i)}\gets f(\textbf{x}^{(i)})$
			\State $\textbf{D}^{(i)}\gets(\textbf{D}^{(i-1)},(\textbf{x}^{(i)\top},y^{(i)}))^\top$
			\EndFor
			\State $(\textbf{x}_{\text{rec}},y_{\text{rec}})\gets\text{Rec}(\textbf{D}^{(I)})$
			\small\Comment{Recommends the best-found optimizer and objective}\normalsize
			\\\Return $(\textbf{x}_{\text{rec}},y_{\text{rec}})$
		\end{algorithmic}
		\label{algo:bo}
	\end{algorithm}
	
	Steps 1-7 of Algorithm \ref{algo:bo} is sometimes referred to as the ``outer'' optimization loop, to distinguish it from the ``inner'' optimization loops at step 2 and step 4 performed at every outer loop iteration. The assumption underpinning the motivation to use BO at all is the fact that step 5, the evaluation of $f$, is very expensive, e.g. a complete FEM simulation -- possibly orders of magnitude costlier than the inner optimization loops. This discrepancy in (computational) cost can be further exploited by expanding step 2 and step 4 appropriately.
	
	
	\subsection{Surrogate model initialization for Bayesian optimization}
	\label{gpi-section}
	New expensive data is sampled during the BO process (step 5 of Algorithm \ref{algo:bo}) during every outer loop iteration. This computationally critical step places substantial importance on selecting the appropriate hyperparameters for BO. In the context of GPR, this latter point equates to the possibility to use any covariance function from a size $K$ tool set $\{\kappa_1,\kappa_2,\ldots,\kappa_K\}$.
	The question of finding the optimal $\kappa$ has previously been posed by in the framework of Bayesian statistics as a so-called Type-II likelihood maximization problem \cite{williams2006gaussian, garnett2023bayesian}. An approach specifically geared towards discovering structure in time series exists. This is undertaken by means of exploring a search space comprised of algebraic compositions from a set of base kernels \cite{duvenaud2013structure}. Despite the aforementioned, there exists no practical implementation of GPR model evaluation and comparison when $D>1$. An informed search methodology is constructed to select a covariance $\hat{\kappa}$ and a corresponding restricted model parameter search space $\hat{\mathcal{T}}\subset\mathcal{T}_{\hat{\kappa}}$ given a set of GPR models arising from optimizing the likelihood in a restricted manner.
	
	Assume that a number of distinct GPR models are constructed based on $\textbf{D}_{\text{train}}$ with a portion of $N_{\text{train}}$ out of the $N$ design rows that populate $\textbf{D}$. By then defining a size $N_{\text{test}}$ design of test experiments withheld from training $\textbf{D}_{\text{test}}:=(\textbf{X}_{\text{test}},\textbf{y}_{\text{test}})$ where $\textbf{X}_{\text{test}}:=(\textbf{x}_{\text{test},1}^\top,\ldots,\textbf{x}_{\text{test},N_{\text{test}}}^\top)^\top$ and $\textbf{y}_{\text{test}}:=(y_{\text{test},1},\ldots,y_{\text{test},N_{\text{test}}})^\top$ as a realization of $f(\textbf{X}_{\text{test}})$. A train-test split commonly used in practice is 20:80, i.e., $N_{\text{test}}/N=1/5$ and $N_{\text{train}}/N=4/5$. This split value is used throughout this manuscript.
	
	Moving forward, a fundamental assumption regarding $f$ needs to be made in order to decide on the quality of an RPD density $\hat{\phi}$. If $f$ is assumed noiseless, i.e. $f(\textbf{X}_{\text{test}})=\textbf{y}_{\text{test}}$ exactly,
	then any deviation between the RPD mean $\hat{\mu}(\textbf{x})$ and the objective evaluation $y=f(\textbf{x})$ for $\textbf{x}\in\textbf{X}_{\text{test}}$ can be interpreted as purely resulting from epistemic uncertainty. In this case, the relative mean squared error (RelMSE) is able to sketch a reasonable picture with regards to the prediction quality of the GP surrogate model. Given a non-constant control vector $\textbf{y}:=(y_1,\ldots,y_M)$ and a prediction vector $\hat{\textbf{y}}:=(\hat{y}_1,\ldots,\hat{y}_M)$, it is defined by
	\begin{equation}
		\mathrm{RelMSE}(\textbf{y}, \hat{\textbf{y}}):=\frac{\mathrm{MSE}(\textbf{y}, \hat{\textbf{y}})}{\mathrm{Var}(\textbf{y})}=\frac{\sum_{j=1}^M(y_j-\hat{y}_j)^2}{\sum_{i=1}^M(y_j-\bar{y})^2}\quad\text{with}\quad\bar{y}:=\frac{1}{M}\sum_{j=1}^My_j.
		\label{RelMSE}
	\end{equation}
	In other words, for a noiseless objective function $f$, the value $\text{RelMSE}(\textbf{y}_{\text{test}},\hat{\mu}(\textbf{X}_{\text{test}}))$ is a precise indicator of the quality of the GPR's RPD. The RelMSE is sometimes also known as the fraction of variance unexplained (FVU), equal to $1-R^2$ where $R^2$ is the coefficient of determination. It is a popular choice of score to measure the (lack of) goodness of fit, especially when comparing regression models applied on different datasets \cite{chicco2021coefficient}.
	
	However, when $f$ is considered noisy, evaluating the quality of $\hat{\phi}$ becomes more complicated.
	This is because $\hat{\mu}$ no longer carries an exact interpolation role through train and test data.
	Instead, $\hat{\mu}(\textbf{x})$ represents the prediction of the mean of $f(\textbf{x})=Y$, a normally distributed random variable. One possible solution is to first let $\textbf{Y}_{\text{test}}:=(\textbf{y}_{\text{test},1}^\top,\ldots,\textbf{y}_{\text{test},N_{\text{test}}}^\top)^\top$ with $\textbf{y}_{\text{test},n}:=(y_{\text{test},n,1},\ldots,y_{\text{test},n,R})^\top$, for each $n\in\{1,\ldots,N_{\text{test}}\}$ and $y_{\text{test},n,r}$ being a realization of $f(\textbf{x}_{\text{test},n})$ for any $r\in\{1,\ldots,R\}$. Then, $\bar{\textbf{y}}_{\text{test}}:=(\bar{\textbf{y}}_{\text{test},1},\ldots,\bar{\textbf{y}}_{\text{test},N_{\text{test}}})^\top$ is an estimator the true mean of $f(\textbf{X}_{\text{test}})$, which indicates $\text{RelMSE}(\bar{\textbf{y}}_{\text{test}},\hat{\mu}(\textbf{X}_{\text{test}}))$ as being a possible error measure in the objective space.
	
	There are major drawbacks to this approach: the RelMSE is being measured between two approximations, while $R$ cannot be large because of the expensive cost of evaluating $f$. In scenarios like these, it will be useful to follow state-of-the-art practice and place RelMSE scoring alongside a probabilistic scoring to judge the quality of $\hat{\phi}$. To this end, define the statistical test log-likelihood (TLL) error score as \cite{deshpande2022you,gelman2014understanding}:
	\begin{equation}
		\text{TLL}(\textbf{D}_{\text{test}},\hat{\phi}):=-\frac{\ln(2\pi)}{2}-\frac{1}{N_{\text{test}}}\sum_{j=1}^{N_{\text{test}}}\frac12\left[\ln(\hat{\sigma}^2(\textbf{x}_j))+\left(\frac{y_j-\hat{\mu}(\textbf{x}_j)}{\hat{\sigma}(\textbf{x}_j)}\right)^2\right],
		\label{eq:tll}
	\end{equation}
	i.e. a sample mean of logarithmic RPD density values evaluated at test outputs. The closer $y_j$ is located to $\hat{\mu}(\textbf{x}_j)$, the higher the value of TLL, which implies higher predictive quality on a probabilistic basis. The TLL also has a global maximum in terms of $\hat{\sigma}^2$, which means that the TLL punishes both overconfident and unconfident predictions. Now, consider the RPD densities $\hat{\phi}$ and $\hat{\phi}'$ arising from two different GPR models. Given the interpretation of the TLL score, one would be inclined to prefer $\hat{\phi}$ over $\hat{\phi}'$ if $\text{TLL}(\textbf{D}_{\text{test}},\hat{\phi})>\text{TLL}(\textbf{D}_{\text{test}},\hat{\phi}')$.
	
	While some authors draw conclusions from their findings based on the TLL alongside predictive mean squared error measurements \cite{duvenaud2013structure}, this is generally speaking not straightforward. Indeed, there exist practical scenarios in which the TLL as a probabilistic quality measure does not correlate with RelMSE as a physical quality measure \cite{deshpande2022you}. In these circumstances, priority should be placed on predictive RelMSE scores, ahead of TLL scores. This is because building a surrogate model for practical engineering applications requires a correspondingly practical measure of error in terms of the (relative) physical units of the objective. This measurement is readily provided by RelMSE, while TLL is a purely statistical score. Concretely, the following is proposed: assume for two RPD densities $\hat{\phi}$ and $\hat{\phi}'$ that 
	$$\text{TLL}(\textbf{D}_{\text{test}},\hat{\phi}')>\text{TLL}(\textbf{D}_{\text{test}},\hat{\phi}),$$
	but simultaneously 
	$$\text{RelMSE}(\textbf{y}_{\text{test}},\hat{\mu}'(\textbf{X}_{\text{test}}))>\text{RelMSE}(\textbf{y}_{\text{test}},\hat{\mu}(\textbf{X}_{\text{test}})).$$
	In this case, preference is assumed for $\hat{\phi}$ over $\hat{\phi}'$ if $R>\text{RelMSE}(\textbf{y}_{\text{test}},\hat{\mu}'(\textbf{X}_{\text{test}}))$ for some threshold $R>0$.
	
	Taking into consideration the high expense of the training data as well as the multimodal nature of many likelihood landscapes, it is prudent to critically investigate solving the likelihood optimization problem for $\hat{\boldsymbol{\theta}}_{\text{MLE}}$ in Equation \eqref{eq:mle}. For a given covariance function $\kappa$ from a set of $K$ covariance functions $K_{\text{set}}:=\{\kappa_1,\ldots,\kappa_K\}$, it might be beneficial for the numerical optimization process to reduce the $T_\kappa$-dimensional search space 
	$$\mathcal{T}_\kappa=P_1\times P_2\times\cdots\times P_{T_\kappa}.$$
	For example, one could define
	$$\mathcal{T}_{\kappa}':=\{\bar{\theta}_1\}\times P_2\times\cdots\times P_{T_\kappa}$$
	for some value $\bar{\theta}_1\in P_1$ and subsequently (numerically) solving for
	\begin{equation}
		\hat{\boldsymbol{\theta}}':=\argmax_{\boldsymbol{\theta}\in{\mathcal{T}_\kappa'}}\ell(\boldsymbol{\theta};\textbf{D})=\argmin_{\boldsymbol{\theta}\in{\mathcal{T}_\kappa'}}\ln(\det(\textbf{K}_{\boldsymbol{\theta}}(\textbf{X})))+\textbf{y}^\top \textbf{K}^{-1}_{\boldsymbol{\theta}}(\textbf{X})\textbf{y}.
		\label{eq:mle-rld}
	\end{equation}
	In this context, $\mathcal{T}_\kappa'$ is called a restricted likelihood domain (RLD).
	
	Given the nonlinear nature of $\ell$, it is possible that the numerical approximation of $\hat{\boldsymbol{\theta}}'$ achieves a higher likelihood than that of $\hat{\boldsymbol{\theta}}_{\text{MLE}}$. If this is the case, there is a quantifiable reason to believe that the training data structure allows for the restriction of the optimization of $\ell$ to $\mathcal{T}_{\kappa}'$, potentially reducing the search space dimensionality for future optimization attempts. 
	
	Of course, there are many other ways to restrict $\mathcal{T}_\kappa$ apart from $\mathcal{T}_\kappa'$. In order to describe the set of RLDs systematically, the following is proposed: 
	
	\begin{itemize}
		\item \textbf{Limited amount of RLDs}. In practice, it is sufficient to consider a small set of popular covariance kernel types: $K_{\text{set}}\doteq\{\kappa_{\text{RBF}},\kappa_{\text{Mat}},\kappa_{\text{RQ}}\}$, i.e. $K=3$. Recall that for these kernels, the following facts hold:
		\begin{align*}
			\boldsymbol{\theta}_{\text{RBF}}&=(c,\lambda,s^2),&T_{\text{RBF}}=3;\\
			\boldsymbol{\theta}_{\text{Mat}}&=(c,\lambda,s^2),&T_{\text{Mat}}=3;\\
			\boldsymbol{\theta}_{\text{RQ}}&=(c,\alpha,\lambda,s^2),&T_{\text{RQ}}=4.
		\end{align*} 
		Given the relatively small dimensionalities of each model parameter space, and without being overly restrictive, it is therefore sufficient to consider only values of $d$ such that $d\le2$. Finally, in order to reduce redundancy, it will be sufficient to consider three nominal values, $\bar{\Theta}_t\doteq\{\bar{\theta}_{t,\text{low}},\bar{\theta}_{t,\text{mid}},\bar{\theta}_{t,\text{high}}\}$, i.e. $V=3$ for all $t$. This readily reduces the size of the search space to $\sum_{k=1}^33\cdot T_{\kappa_k}+9\cdot\binom{T_{\kappa_k}}{2}$.
		\item \textbf{Structured search}. A grid search policy with breadth-first focus is proposed to define a sequence of RLDs $\mathcal{T}_1,\mathcal{T}_2,\ldots$ to solve the restricted likelihood optimization problem.
		\begin{itemize}
			\item If a previous RLD search has been performed at an earlier BO iteration, with result $\mathcal{T}_{\kappa}(\bar{\boldsymbol{\theta}}_F)$, then use the mixed RelMSE / TLL performance measure, to assess the quality of RPD $\hat{\phi}$ corresponding to the numerically optimized model parameter vector
			$$\hat{\boldsymbol{\theta}}=\argmin_{\boldsymbol{\theta}\in{\mathcal{T}_{\kappa}(\bar{\boldsymbol{\theta}}_F)}}\ln(\det(\textbf{K}_{\boldsymbol{\theta}}(\textbf{X})))+\textbf{y}^\top \textbf{K}^{-1}_{\boldsymbol{\theta}}(\textbf{X})\textbf{y}.$$
			If the RPD quality is sufficient, then the search is terminated. In all other cases, continue the search with the next step.
			\item The RLD search starts with regular, state-of-the-art \textit{unrestricted} likelihood optimization as defined in Equation \eqref{eq:mle}. This means:
			$$\mathcal{T}_1=\mathcal{T}_{\text{RBF}},\quad\mathcal{T}_2=\mathcal{T}_{\text{Mat}},\quad\mathcal{T}_3=\mathcal{T}_{\text{RQ}}.$$
			\item For any subsequent RLD, the search algorithm will similarly cycle through the set of covariance kernels for each $d$ in increasing value. For each $d$, all possible index sets $F$ are considered, and correspondingly, all possible $\bar{\boldsymbol{\theta}}_F\in\bar{\Theta}_F$. As an explicit example with the setting described previously,
			\begin{align*}
				\mathcal{T}_4&=\mathcal{T}_{\text{RBF}}(\bar{c}_{\text{low}}),\quad
				\mathcal{T}_5=\mathcal{T}_{\text{RBF}}(\bar{c}_{\text{mid}}),\quad
				\mathcal{T}_6=\mathcal{T}_{\text{RBF}}(\bar{c}_{\text{high}}),\\
				\mathcal{T}_7&=\mathcal{T}_{\text{RBF}}(\bar{\lambda}_{\text{low}}),\quad
				\mathcal{T}_8=\mathcal{T}_{\text{RBF}}(\bar{\lambda}_{\text{mid}}),\quad
				\mathcal{T}_9=\mathcal{T}_{\text{RBF}}(\bar{\lambda}_{\text{high}}),\\
				&\ldots,\\
				\mathcal{T}_{13}&=\mathcal{T}_{\text{Mat}}(\bar{c}_{\text{low}}),\quad
				\mathcal{T}_{14}=\mathcal{T}_{\text{Mat}}(\bar{c}_{\text{mid}}),\quad
				\mathcal{T}_{14}=\mathcal{T}_{\text{Mat}}(\bar{c}_{\text{mid}})\\
				&\ldots,\\
				\mathcal{T}_{31}&=\mathcal{T}_{\text{RBF}}(\bar{c}_{\text{low}},\bar{\lambda}_{\text{low}}),\quad
				\mathcal{T}_{32}=\mathcal{T}_{\text{RBF}}(\bar{c}_{\text{mid}},\bar{\lambda}_{\text{low}}),\quad\ldots
			\end{align*}
		\end{itemize}
		Apart from this structured grid search, other search methods exist which are typically used to search the a space of (hyper)parameters. These include (naïve) grid search, quasi-random search and tree-structured Parzen estimators \cite{watanabe2023tree}. Off-the-shelf packages facilitate a practical implementation of these methods fo supervised machine learning models in general, such as \texttt{SMAC} \cite{hutter2011sequential}, \texttt{Hyperopt} \cite{bergstra2015hyperopt} and \texttt{Optuna} \cite{optuna_2019}.
	\end{itemize}
	
	The breadth-first grid search process of GP model selection or initialization (GPi) is summarized in Algorithm \ref{algo:gpi}. 
	
	\begin{algorithm}[H]
		\caption{Gaussian process model selection / initialization (GPi)}
		\begin{algorithmic}[1]
			\Require{Design of train experiments $\textbf{D}$, design of test experiments $\textbf{D}_{\text{test}}$, set of covariance kernels $\{\kappa_{\text{RBF}},\kappa_{\text{Mat}},\kappa_{\text{RQ}}\}$, collection of sets of parameter indices to fix $\{\mathcal{F}_0=\varnothing,\mathcal{F}_1,\mathcal{F}_2\}$, nominal parameter fixture values $\{\{\bar{\theta}_{t,\text{low}},\bar{\theta}_{t,\text{mid}},\bar{\theta}_{t,\text{high}}\}:t\in\{1,\ldots,D\}\}$, trial threshold $Q$, RelMSE threshold $R$}
			\State $\text{RelMSE}_*\gets+\infty$
			\State $\text{TLL}_*\gets-\infty$
			\State $\hat{\kappa},\ \hat{\mathcal{T}},\ \hat{\boldsymbol{\theta}}\gets\varnothing$
			\State $q,k\gets0$
			\For{$d\in\{0,1,2\}$}
			\For{$\kappa\in K_{\text{set}}$}
			\For{$F\in\mathcal{F}_d$}
			\State $\bar{\Theta}_F\gets\prod_{t\in F}\{\bar{\theta}_{t,\text{low}},\bar{\theta}_{t,\text{mid}},\bar{\theta}_{t,\text{high}}\}$
			\For{$\bar{\boldsymbol{\theta}}_F\in\bar{\Theta}_F$}
			\footnotesize\State $\hat{\boldsymbol{\theta}}_{\text{MLE}}\gets\argmin_{\boldsymbol{\theta}\in\mathcal{T}_\kappa(\bar{\boldsymbol{\theta}}_F)}\ln(\det(\textbf{K}_{\boldsymbol{\theta}}(\textbf{X})))+\textbf{y}^{\top} \textbf{K}^{-1}_{\boldsymbol{\theta}}(\textbf{X}) \textbf{y}$
			\Comment{Eq. \eqref{eq:mle}, \eqref{eq:mle-rld}}\normalsize
			\State $q\gets q+1$
			\State $\hat{\phi}\gets\hat{\boldsymbol{\theta}}_{\text{MLE}}$
			\footnotesize\State $\rho:=\text{RelMSE}(\textbf{y}_{\text{test}},\hat{\mu}(\textbf{X}_{\text{test}})),\ T:=\text{TLL}(\textbf{D}_{\text{test}},\hat{\phi})\gets\hat{\phi}$
			\Comment{Eq. \eqref{RelMSE}, \eqref{eq:tll}}\normalsize
			\If{
				$\rho<\text{RelMSE}_*$ or $R>\rho>\text{RelMSE}_*$ and $T>\text{TLL}_*$
			}
			\State $\text{RelMSE}_*\gets\rho$
			\State $\text{TLL}_*\gets T$
			\State $\hat{\kappa},\ \hat{\mathcal{T}},\ \hat{\boldsymbol{\theta}}\gets\kappa,\ \mathcal{T}_\kappa(\bar{\boldsymbol{\theta}}_F),\ \hat{\boldsymbol{\theta}}_{\text{MLE}}$
			\EndIf
			\If{$\text{RelMSE}_*<0.05$ or $q\ge Q$}
			\Return $\hat{\kappa},\ \hat{\mathcal{T}},\ \hat{\boldsymbol{\theta}}$
			\EndIf
			\EndFor
			\EndFor
			\EndFor	
			\EndFor\\
			\Return $\hat{\kappa},\ \hat{\mathcal{T}},\ \hat{\boldsymbol{\theta}}$
		\end{algorithmic}
		\label{algo:gpi}
	\end{algorithm}
	
	\subsection{Adaptive candidate design selection for BO}
	\label{bo-iada-section}
	
	A reasonable next step in this discussion is exploring the possibilities to automate the selection of $\alpha$. In order to do this, it is reasonable to exploit the assumption that the optimization of an acquisition function $\alpha$ is computationally much cheaper than optimizing $f$. Thus, given a toolbox of acquisition functions $\alpha_1,\alpha_2,\ldots$, it is possible to optimize each of them efficiently and yield design candidates $\textbf{x}_1^{(i)},\textbf{x}_2^{(i)},\ldots$, with outer iteration step $i$. Then, in order to find the best possible candidate design in terms of optimizing $f$, each of these designs should ideally be evaluated at every iteration. However, since $f$ is an expensive objective, this is not always feasible. 
	
	As a direct comparison between objective evaluations $f(\textbf{x}_1^{(i)}),f(\textbf{x}_2^{(i)}),\ldots$ is impractical due to the bottleneck of computational expense, one might turn to comparing acquisition functions instead. However, this method has a number of drawbacks. 
	
	Firstly, it should be emphasized that each acquisition landscape represents a belief model given the available data, and might operate on a different scale of magnitude compared to another acquisition. For example, the range of $\alpha_{\text{PI}}$ is the unit interval $(0,1)$, while the range of $\alpha_{\text{UCB}}$ is $\mathbb{R}$. This fact rules out any direct comparisons between acquisition function values.
	
	The dilemma of choosing between different acquisition recommendations is further exacerbated by the fact that it is difficult to discern the quality of the different belief models and their recommendations, as they are each based on a different statistical metric \cite{garnett2023bayesian}. In other words, it cannot be generally stated that, or when, one acquisition function yields better design suggestions than another one. This lack of prior knowledge inspires an application of selection methodologies originating from reinforcement learning, in particular $k$-armed bandits \cite{barto1997reinforcement}. While the theory of $k$-armed bandits provides a reward-based strategy towards candidate selection, there is no accompanying measure of reward -- stochastic or deterministic -- that is both intuitive and easily described or modelled. For this reason, there is not much motivation to employ (partly) deterministic selection schemes such as the ($\varepsilon$-)greedy algorithm. Facing these difficulties, alternative criteria need to be devised in order to select a candidate.
	
	Suppose that $\mathcal{A}:=\{\alpha_1,\ldots,\alpha_A\}$ is the set of possible acquisition functions in a given toolbox. Each acquisition $\alpha_a:[0,1]^D\to\mathbb{R}$, where $a\in\{1,\ldots,A\}$, can be optimized to create a matrix of candidate designs $\textbf{X}^{(i)}_\text{cand}:=(\textbf{x}^{(i)\top}_1,\ldots,\textbf{x}^{(i)\top}_A)^\top$ at iteration $i\in\{1,\ldots,I\}$, where
	\begin{equation}
		\textbf{x}^{(i)}_a:=\argmax_{\textbf{x}\in[0,1]^D}\alpha_a(\textbf{x};\hat{\phi}^{(i)}).
	\end{equation}
	
	A selection strategy, generically denoted by ``Sel'', outputs $\textbf{x}^{(i)}_a$ for some $a\in\{1,\ldots,A\}$ as a response to the candidate design matrix $\textbf{X}^{(i)}_\text{cand}$ and all available data $\textbf{D}^{(i-1)}$.
	
	One possible strategy consists of selecting $\textbf{x}^{(i)}_a$ randomly. Out of the possible random strategies, uniform random sampling is the most straightforward:
	
	\begin{equation}
		\text{Sel}_{\mathcal{U}}(\textbf{X}^{(i)}_{\text{cand}},\textbf{D}^{(i-1)})=\text{Sel}_{\mathcal{U}}(\textbf{X}^{(i)}_{\text{cand}}):=\textbf{x}^{(i)}_a,\quad a\gets\mathcal{U}_{\{1,\ldots,A\}}
		\label{eq:sel-u}
	\end{equation}
	It should be noted that the selection strategy in Equation \eqref{eq:sel-u} does not actually depend on any of the previously available data $\textbf{D}^{(i-1)}$. However, inspired by the approach taken by solving the $k$-armed bandits problem, a (fully stochastic) selection method will be described which does make use of the available data at optimization iteration $i$.
	
	Let $N$ be a positive integer and let $\textbf{p}=(p_1,\ldots,p_N)^\top$ be a probability vector, i.e. $0\le p_n\le1$ for all $n\in\{1,\ldots,N\}$ and $\sum_{n=1}^Np_n=1$. Let $\text{Cat}(N,\textbf{p})$ denote the categorical probability distribution supported on $\{1,\ldots,N\}$, defined by probability mass function $P(C=n)=p_n$ for any $n\in\{1,\ldots,N\}$ if $C\sim\text{Cat}(N,\textbf{p})$. 
	
	Define $p^{(1)}_a:=1/A$ for all $a\in\{1,\ldots,A\}$. For $i>1$, let $a^{(i-1)}$ be the selected value for $a$ at iteration $i-1$. Then, define the categorical probability vector $\textbf{p}^{(i)}=\textbf{p}^{(i)}(\textbf{D}^{(i-1)}):=(p^{(i)}_1,\ldots,p^{(i)}_A)^\top$ recursively as follows:
	\begin{multline}
		p^{(i)}_a:=\frac{n^{(i)}_a}{N^{(i)}}\quad\text{with}\quad 
		n^{(i)}_a:=\left\{
		\begin{array}{ll}
			n^{(i-1)}_a+1&\text{if }a=a^{(i-1)}\text{ and }y^{(i-1)}=\min\textbf{y}^{(i-1)},\\
			n^{(i-1)}_a&\text{otherwise},
		\end{array}
		\right.\\ 
		\text{where}\quad n_a^{(1)}:=1\quad\text{and}\quad N^{(i)}:=\sum_{a=1}^An^{(i)}_a.
	\end{multline}
	Then, the categorical (also called multinomial) candidate design selection strategy can be formulated as follows:
	\begin{equation}
		\text{Sel}_{\text{Cat}}(\textbf{X}^{(i)}_{\text{cand}},\textbf{D}^{(i-1)}):=\textbf{x}^{(i)}_a,\quad a=a^{(i)}\gets\text{Cat}(A,\textbf{p}^{(i)}(\textbf{D}^{(i-1)})).
		\label{sel-mn}
	\end{equation}
	In other words, $\text{Sel}_{\text{Cat}}$ will assign a larger probability to select acquisition function $\alpha_a$ if it was able to locate the incumbent optimum at the previous iteration, whereas the probability mass distribution over the set of available acquisition functions will remain the same otherwise. This selection strategy is inspired by the \texttt{Dragonfly} implementation \cite{k2019tuning}, the authors of which take a similar approach for selecting subsequent candidate designs and regression models as the outer BO steps progress.
	
	The availability of previously sampled data $\textbf{D}^{(i-1)}$ at iteration $i$ can be exploited further when devising candidate design selection strategies for objective evaluation. While categorical selection defined in Equation \eqref{sel-mn} only makes use of $\textbf{y}^{(i-1)}$, the same selection methodology -- and indeed uniform random selection, Equation \eqref{eq:sel-u} -- can be expanded based on $\textbf{X}^{(i-1)}$. 
	
	BO needs to employ a careful trade-off between exploration and exploitation of the design space, especially applied to expensive problems. It will precisely be inefficient to sample two very similar designs twice, without exploring the design space first, even if one or both of them have been suggested by optimizing an acquisition function. It is therefore in order to encourage exploration when necessary, but still allow exploitation of promising design candidates as $i$ approaches $I$. Fortunately, because there is a host of design candidates $\textbf{X}_{\text{cand}}^{(i)}$ to choose from, those candidates which are overly exploitative can be ruled out: a candidate design can be rejected based on its proximity to $\textbf{X}^{(i-1)}$. 
	
\begin{figure}[H]
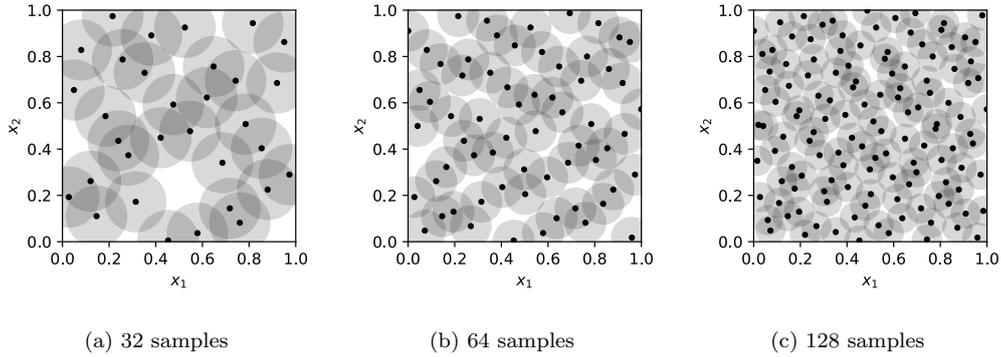

		\centering
		\begin{subfigure}{.3\textwidth}
			\includesvg[width=\textwidth]{img/diagrams/Sobol_N32_exploit.svg}
			\caption{32 samples}
		\end{subfigure}
		\begin{subfigure}{.3\textwidth}
			\includesvg[width=\textwidth]{img/diagrams/Sobol_N64_exploit.svg}
			\caption{64 samples}
		\end{subfigure}
		\begin{subfigure}{.3\textwidth}
			\includesvg[width=\textwidth]{img/diagrams/Sobol_N128_exploit.svg}
			\caption{128 samples}
		\end{subfigure}
		\caption{Two-dimensional design space samples (black dots) with disks (gray) of radius MMD.}
		\label{fig:mmd-circles}
	\end{figure}

	In order to make this notion of clustering concrete, the following is proposed: for any $N$ design parameter vectors summarized in a matrix $\textbf{U}:=(\textbf{u}_1^\top,\ldots,\textbf{u}_N^\top)^\top$, let $\delta_1,\ldots,\delta_N$ be the minimum Euclidean distances between $\textbf{U}$ and itself, defined as follows for any $j\in\{1,\ldots,N\}$:
	\begin{equation}
		\delta_j:=\min_{\substack{j'\in\{1,\ldots,N\}\\j'\ne j}}\|\textbf{u}_{j'}-\textbf{u}_j\|.
	\end{equation}
	Subsequently, define the median minimum distance (MMD) of $\textbf{U}$ as
	\begin{equation}
		\mathrm{MMD}(\textbf{U})=\textrm{median}\{\delta_1,\ldots,\delta_N\}.
		\label{mmd}
	\end{equation}
	An illustration of the MMD on different sample sets is shown in Figure \ref{fig:mmd-circles}.
	
	Next, define
	\begin{equation}
		d_{\min}(\textbf{x}^{(i)}_a,\textbf{X}^{(i-1)}):=\min_{j\in\{1,\ldots,N+(i-1)\}}\|\textbf{x}^{(i)}_a-\textbf{x}^{(i-1)}_j\|
	\end{equation}
	for $a\in\{1,\ldots,A\}$ as the minimal Euclidean distance between $\textbf{x}^{(i)}_a$ and any design row in $\textbf{X}^{(i-1)}$.
	
	It is now possible to calculate $\text{MMD}(\textbf{X}^{(i-1)})$, the median minimum distance of the design matrix at iteration $i-1$, and compare this value to $d_{\min}(\textbf{x}^{(i)}_a,\textbf{X}^{(i-1)})$. If $\text{MMD}(\textbf{X}^{(i-1)})\gg d_{\min}(\textbf{x}^{(i)}_a,\textbf{X}^{(i-1)})$, then $\textbf{x}^{(i)}_a$ is a relatively exploitative design, while $\text{MMD}(\textbf{X}^{(i-1)})\ll d_{\min}(\textbf{x}^{(i)}_a,\textbf{X}^{(i-1)})$ indicates exploration by $\textbf{x}_a^{(i)}$.
	
	Thus, introducing the exploitation score (ES) of $\textbf{x}^{(i)}_a$ with respect to $\textbf{X}^{(i-1)}$ as
	\begin{equation}
		\text{ES}(\textbf{x}^{(i)}_a,\textbf{X}^{(i-1)}):=\ln\left(\frac{\text{MMD}(\textbf{X}^{(i-1)})}{d_{\min}(\textbf{x}^{(i)}_a,\textbf{X}^{(i-1)})}\right),
	\end{equation}
	it can be decided to refrain from evaluating the expensive objective $f$ at $\textbf{x}^{(i)}_a$ if $\text{ES}(\textbf{x}^{(i)}_a,\textbf{X}^{(i-1)})>t^{(i)}$ for some threshold value $t^{(i)}\in\mathbb{R}$. The explicit choice to make this threshold value depend on the outer loop iteration $i$ stems from the desire to encourage exploration when $i$ is small, yet allow exploitation when $i$ is large. In general, $t^{(i)}$ is therefore programmed to decrease with respect to $i$. It should be noted that the logarithmic nature of ES can conforms with the desirable property that a difference in exploitation score is proportional to the difference in \textit{magnitude} of the distance ratio.
	
	Figure \ref{fig:mmd-circles} showcases the intuitive notion that the MMD decreases as the number of samples increases. This implies a natural yet adaptive notion of which candidate designs are exploitative or exploratory, given a similar ES threshold parameter $t^{(i)}$: a candidate near a previously evaluated design sample in the 32 samples case is less exploitative in a similar 128 samples scenario.
	
	The samples used in Figure \ref{fig:mmd-circles} are so-called Sobol' samples \cite{sobol2001global}, which are based on a low-discrepancy quasi-random Sobol' sequence. The qualities of these samples are such that they are fully predictable (unlike random samples \cite{hastings1970monte}) and can be easily extended to include an arbitrary number of samples (unlike Latin hypercube samples \cite{mckay2000comparison}). Furthermore, Sobol' samples can be used to obtain Saltelli samples \cite{saltelli2010variance}, which is the core component of performing variance-based sensitivity analysis.
	
	\subsection{Adaptive Bayesian optimization}
	By combining the ideas from surrogate model initialization and adaptive candidate selection with the BO algorithm (Algorithm \ref{algo:bo}), a set of extensions can be devised.
	
	\begin{algorithm}[H]
		\caption{Bayesian optimization with Gaussian process initialization and input-adaptive candidate design selection (BO-GPi-iAda, adaptive BO)}
		\begin{algorithmic}[1]
			\Require{Design of train experiments $\textbf{D}^{(0)}$, design of test experiments $\textbf{D}_{\text{test}}$, set of covariance kernels $K_{\text{set}}=\{\kappa_{\text{RBF}},\kappa_{\text{Mat}},\kappa_{\text{RQ}}\}$, collection of sets of parameter indices to fix $\mathcal{F}=\{\mathcal{F}_0=\varnothing,\mathcal{F}_1,\mathcal{F}_2\}$, nominal parameter fixture values $\bar{\Theta}=\{\{\bar{\theta}_{t,\text{low}},\bar{\theta}_{t,\text{mid}},\bar{\theta}_{t,\text{high}}\}:t\in\{1,\ldots,D\}\}$, trial threshold $Q$, set of acquisition functions $\mathcal{A}$, number of iterations $I$, GPR RelMSE threshold $R$, GP initialization condition $C$, design candidate selection strategy Sel, exploitation score threshold $\textbf{t}$}
			\For{$i=1,\ldots,I$}    
			\If{$C(i)$ or $i=1$}
			\State $\hat{\kappa},\ \hat{\mathcal{T}},\ \hat{\boldsymbol{\theta}}_{\text{MLE}}^{(i)}\gets\text{GPi}(\textbf{D}^{(i-1)},\textbf{D}_{\text{test}},K_{\text{set}},\mathcal{F},\bar{\Theta},Q,R)$
			\Comment{Algorithm \ref{algo:gpi}}
			\Else
			\footnotesize\State$\hat{\boldsymbol{\theta}}_{\text{MLE}}^{(i)}\gets\argmin_{\boldsymbol{\theta}\in\hat{\mathcal{T}}}\ln(\det(\hat{\textbf{K}}_{\boldsymbol{\theta}}(\textbf{X}^{(i-1)})))+\textbf{y}^{(i-1)\top} \hat{\textbf{K}}^{-1}_{\boldsymbol{\theta}}(\textbf{X}^{(i-1)}) \textbf{y}^{(i-1)}$
			\Comment{Eq. \eqref{eq:mle}}\normalsize
			\EndIf
			\State $\hat{\phi}^{(i)}\gets\hat{\boldsymbol{\theta}}_{\text{MLE}}^{(i)}$
			\State $\textbf{X}^{(i)}_{\text{cand}}\gets(\argmax_{\textbf{x}\in[0,1]^D}\alpha_a(\textbf{x};\hat{\phi}^{(i)}))_{a=1,\ldots,A}$
			\State $\textbf{X}^{(i)}_{\text{cand}}\gets(\textbf{x}^{(i)}_a\in\textbf{X}^{(i)}_{\text{cand}}:\text{ES}(\textbf{x}^{(i)}_a,\textbf{X}^{(i-1)})\le t^{(i)}$)\scriptsize\Comment{Skip if no candidate satisfies ES threshold.}\normalsize
			\State $\textbf{x}^{(i)}\gets\text{Sel}(\textbf{X}^{(i)}_{\text{cand}},\textbf{D}^{(i-1)})$
			\State $y^{(i)}\gets f(\textbf{x}^{(i)})$
			\State $\textbf{D}^{(i)}\gets(\textbf{D}^{(i-1)},(\textbf{x}^{(i)\top},y^{(i)}))^\top$
			\EndFor
			\State $(\textbf{x}_{\text{rec}},y_{\text{rec}})\gets\text{Rec}(\textbf{D}^{(I)})$
			\Comment{Recommends the best-found optimizer and objective}
			\\\Return $(\textbf{x}_{\text{rec}},y_{\text{rec}})$
		\end{algorithmic}
		\label{algo:bo-gpi-iada}
	\end{algorithm}
	
\begin{figure}[H]
		\centering
		\includesvg[width=.75\textwidth]{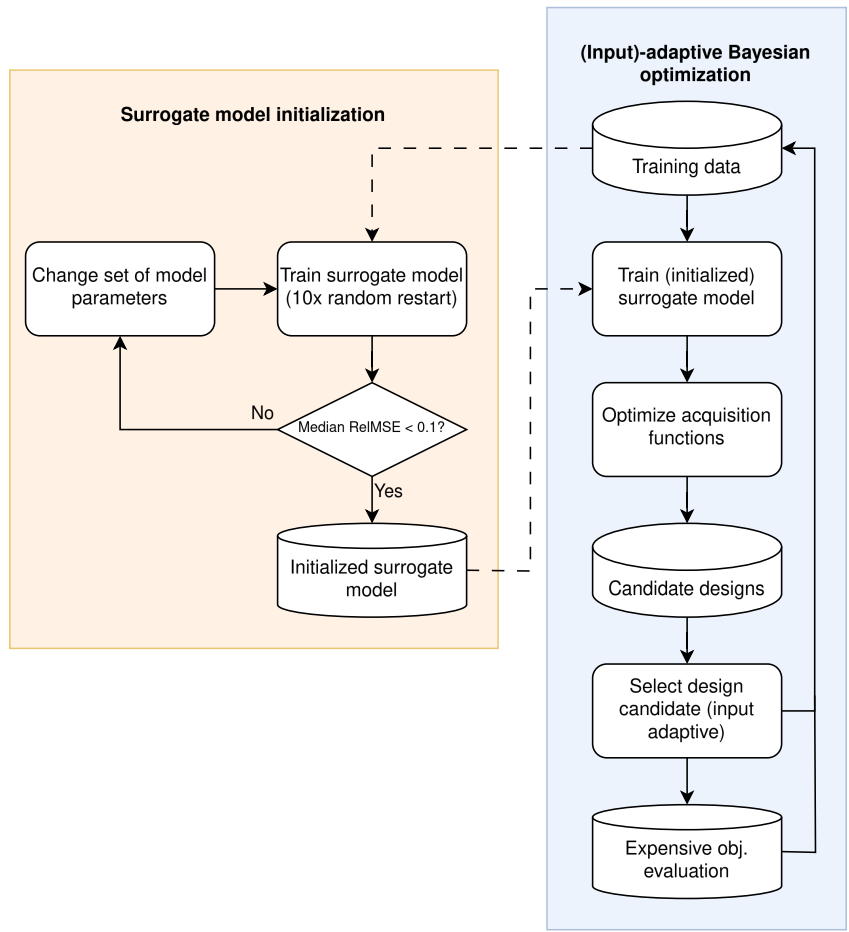}
		\caption{Flowchart diagram of BO-GPi-(i)Ada, Algorithm \ref{algo:bo-gpi-iada}. The dashed arrows represent connections that are only active when $C(i)$ is true.}
		\label{fig:bo-gpi-iada-diagram}
	\end{figure}

	BO-GPi-Ada is analogously defined by removing the exploitation score threshold step 11 of BO-GPi-iAda. Similarly, BO-(i)Ada opts out on the GPi portion (steps 2-5) and performs the maximum likelihood estimation in the same way as is done in regular BO. Finally, BO-GPi hinges on one single acquisition function --- similar to BO --- while retaining the covariance kernel selection steps. See Figure \ref{fig:bo-gpi-iada-diagram} for a schematic overview of BO-GPi-iAda.
	
	Given the newly introduced algorithms, they are now compared to standard BO under a similar optimization budget constraint. To this end, two synthetic objective functions of different landscape qualities are introduced for the optimization algorithms to minimize. See Table \ref{tab:synthetic-functions} for a description of these synthetic objectives.
	
	\begin{table}[H]
		\centering
		\caption{Selected synthetic objective functions to benchmark the BO schemes.}
		\footnotesize
		\begin{tabular}{lllll}
			Function name & Formula                             & Unscaled domain & Multimodal & Global minimum                                \\ \hline
			AlpineN2      & $-\prod_{d=1}^D\sqrt{x_d}\sin(x_d)$ & $[0, 10]^D$     & Yes        & $x_d\approx7.91$ \\ \hline
			Sphere        & $\sum_{d=1}^Dx_d^2$                 & $[-5, 5]^D$     & No         & $x_d=0$         \\ \hline
		\end{tabular}
		\label{tab:synthetic-functions}
	\end{table}
	
\begin{figure}[H]
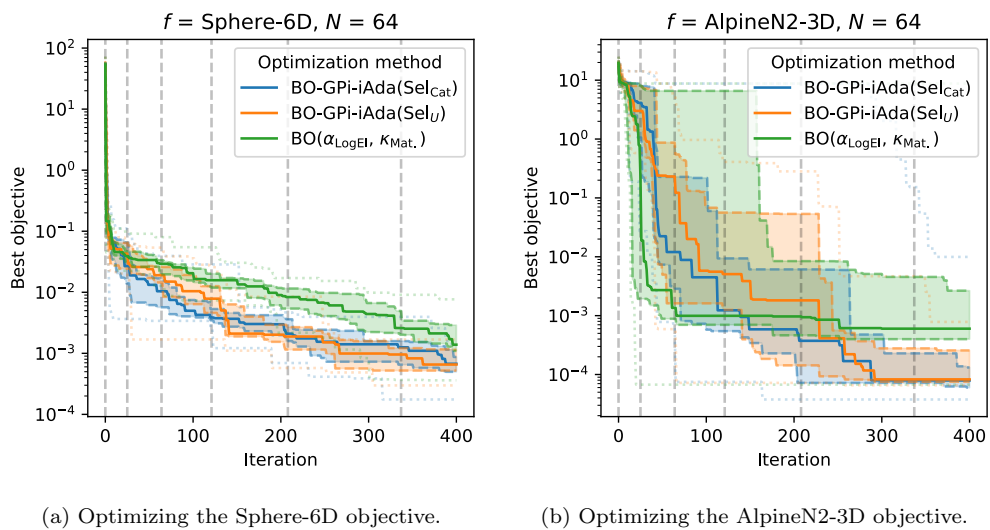

		\centering
		\begin{subfigure}{.45\textwidth}
			\includesvg[width=\textwidth]{img/obj_sfbo_vs_gpi_adaptive_figure_6_Sphere_1.0_64_128_Sogpr_1.0.svg}
			\caption{Optimizing the Sphere-6D objective.}
			\label{fig:gpi-iada-Sphere-6D}
		\end{subfigure}
		\begin{subfigure}{.45\textwidth}
			\includesvg[width=\textwidth]{img/obj_sfbo_vs_gpi_adaptive_figure_3_AlpineN2_1.0_64_128_Sogpr_1.0.svg}
			\caption{Optimizing the AlpineN2-3D objective.}
			\label{fig:gpi-iada-AlpineN2-3D}
		\end{subfigure}
		\caption{Five-number summary of BO (Algorithm \ref{algo:bo}) and BO-GPi-iAda optimization histories (Algorithm \ref{algo:bo-gpi-iada}). A comparison between BO, with the Matérn covariance kernel and the logarithmic Expected Improvement acquisition function as chosen hyperparameters, and BO-GPi-iAda with two different candidate selection methods. The incumbent minimal objective value is plotted against the iteration. The vertical gray dashed lines indicate the iterations at which GPi takes place.}
		\label{fig:gpi-iada-performance}
	\end{figure}

	Figure \ref{fig:gpi-iada-performance} shows the performance of BO-GPi-iAda compared to standard BO when optimizing the six-dimensional Sphere function (Sphere-6D) and the three-dimensional AlpineN2 function (AlpineN2-3D) over 64 initial Sobol' samples and 400 outer loop iterations. In the adaptive BO scenarios, 20\% of the samples, rounded down, are held back for testing during GPi. As an example, this corresponds to 12 design samples when GPi is first performed, at the first adaptive BO iteration.
	
	The core motivation behind using adaptive hyperparameters with BO is to at least perform better than the worst-case standard BO scenario. Let $\Omega:=\{O_1,\ldots,O_M\}$ be a collection of $M$ (reference) optimizers. Next, let
	\begin{equation}
		\textbf{w}_k(\Omega):=(\max\{q^{(i)}_k(O_1),\ldots,q^{(i)}_k(O_M)\})_{i=1,\ldots,I},
	\end{equation} 
	be the $k$-th quartile worst-case aggregate across $\Omega$. Then, for a collection of $M'$ optimizers $\Omega':=\{O_1,\ldots,O_{M'}\}$ which are to be compared to $\Omega$, define
	\begin{equation}
		\text{WCRI}_k(\Omega,\Omega'):=\text{median}\left(\frac{\textbf{w}_k(\Omega)-\textbf{w}_k(\Omega')}{\textbf{w}_k(\Omega)}\right)=1-\text{median}\left(\frac{\textbf{w}_k(\Omega')}{\textbf{w}_k(\Omega)}\right)
		\label{eq:WCRI-double-collective}
	\end{equation}
	as a worst-case variant of the relative improvement in the $k$-th quartile.
	
	Given this new measure between two sets of optimizers, the optimization runs of the synthetic objectives that gave rise to Figure \ref{fig:gpi-iada-performance} are reconsidered and expanded upon. The results are presented in Table \ref{tab:wcri-synthetic}.
	
	\begin{table}[H]
		\centering
		\caption{Table of WCRI values of adaptive Bayesian optimizers compared to a standard BO reference with towards the minimization of the indicated synthetic objective functions.}
		\footnotesize
		\begin{tabular}{|l|l|l|l|r|r|r|r|r|}
			\toprule
			&  &  &  & \multicolumn{5}{r}{Relative improvement per quartile (\%)} \\
			&  &  &  & $Q_0$ & $Q_1$ & $Q_2$ & $Q_3$ & $Q_4$ \\
			Objective & Model init. & Adaptivity & Sel &  &  &  &  &  \\
			\midrule
			\multirow[c]{10}{*}{Sphere-6D} & \multirow[c]{5}{*}{No GPi} & No Ada &  & {\cellcolor[HTML]{FEFFBE}} \color[HTML]{000000} 0.0 & {\cellcolor[HTML]{FEFFBE}} \color[HTML]{000000} 0.0 & {\cellcolor[HTML]{FEFFBE}} \color[HTML]{000000} 0.0 & {\cellcolor[HTML]{FEFFBE}} \color[HTML]{000000} 0.0 & {\cellcolor[HTML]{FEFFBE}} \color[HTML]{000000} 0.0 \\
			\cline{3-9} \cline{4-9}
			&  & \multirow[c]{2}{*}{Ada} & $\mathrm{Sel}_{\mathrm{Cat}}$ & {\cellcolor[HTML]{2AA054}} \color[HTML]{F1F1F1} 75.6 & {\cellcolor[HTML]{36A657}} \color[HTML]{F1F1F1} 72.2 & {\cellcolor[HTML]{60BA62}} \color[HTML]{F1F1F1} 61.5 & {\cellcolor[HTML]{A2D76A}} \color[HTML]{000000} 41.2 & {\cellcolor[HTML]{B3DF72}} \color[HTML]{000000} 34.4 \\
			\cline{4-9}
			&  &  & $\mathrm{Sel}_{U}$ & {\cellcolor[HTML]{1E9A51}} \color[HTML]{F1F1F1} 78.8 & {\cellcolor[HTML]{4EB15D}} \color[HTML]{F1F1F1} 65.7 & {\cellcolor[HTML]{75C465}} \color[HTML]{000000} 55.4 & {\cellcolor[HTML]{7AC665}} \color[HTML]{000000} 53.8 & {\cellcolor[HTML]{D3EC87}} \color[HTML]{000000} 22.2 \\
			\cline{3-9} \cline{4-9}
			&  & \multirow[c]{2}{*}{iAda} & $\mathrm{Sel}_{\mathrm{Cat}}$ & {\cellcolor[HTML]{16914D}} \color[HTML]{F1F1F1} 82.4 & {\cellcolor[HTML]{128A49}} \color[HTML]{F1F1F1} 85.9 & {\cellcolor[HTML]{39A758}} \color[HTML]{F1F1F1} 71.5 & {\cellcolor[HTML]{4EB15D}} \color[HTML]{F1F1F1} 65.9 & {\cellcolor[HTML]{93D168}} \color[HTML]{000000} 46.1 \\
			\cline{4-9}
			&  &  & $\mathrm{Sel}_{U}$ & {\cellcolor[HTML]{148E4B}} \color[HTML]{F1F1F1} 84.2 & {\cellcolor[HTML]{15904C}} \color[HTML]{F1F1F1} 83.4 & {\cellcolor[HTML]{2AA054}} \color[HTML]{F1F1F1} 75.1 & {\cellcolor[HTML]{48AE5C}} \color[HTML]{F1F1F1} 67.9 & {\cellcolor[HTML]{78C565}} \color[HTML]{000000} 54.2 \\
			\cline{2-9} \cline{3-9} \cline{4-9}
			& \multirow[c]{5}{*}{GPi} & No Ada &  & {\cellcolor[HTML]{0E8245}} \color[HTML]{F1F1F1} 88.5 & {\cellcolor[HTML]{108647}} \color[HTML]{F1F1F1} 86.8 & {\cellcolor[HTML]{18954F}} \color[HTML]{F1F1F1} 80.9 & {\cellcolor[HTML]{199750}} \color[HTML]{F1F1F1} 79.8 & {\cellcolor[HTML]{2AA054}} \color[HTML]{F1F1F1} 75.5 \\
			\cline{3-9} \cline{4-9}
			&  & \multirow[c]{2}{*}{Ada} & $\mathrm{Sel}_{\mathrm{Cat}}$ & {\cellcolor[HTML]{08773F}} \color[HTML]{F1F1F1} 93.4 & {\cellcolor[HTML]{06733D}} \color[HTML]{F1F1F1} 94.5 & {\cellcolor[HTML]{0B7D42}} \color[HTML]{F1F1F1} 90.7 & {\cellcolor[HTML]{0E8245}} \color[HTML]{F1F1F1} 88.6 & {\cellcolor[HTML]{0E8245}} \color[HTML]{F1F1F1} 89.0 \\
			\cline{4-9}
			&  &  & $\mathrm{Sel}_{U}$ & {\cellcolor[HTML]{06733D}} \color[HTML]{F1F1F1} 95.1 & {\cellcolor[HTML]{0D8044}} \color[HTML]{F1F1F1} 89.1 & {\cellcolor[HTML]{118848}} \color[HTML]{F1F1F1} 86.6 & {\cellcolor[HTML]{18954F}} \color[HTML]{F1F1F1} 80.6 & {\cellcolor[HTML]{16914D}} \color[HTML]{F1F1F1} 82.5 \\
			\cline{3-9} \cline{4-9}
			&  & \multirow[c]{2}{*}{iAda} & $\mathrm{Sel}_{\mathrm{Cat}}$ & {\cellcolor[HTML]{05713C}} \color[HTML]{F1F1F1} 96.1 & {\cellcolor[HTML]{07753E}} \color[HTML]{F1F1F1} 94.4 & {\cellcolor[HTML]{07753E}} \color[HTML]{F1F1F1} 93.9 & {\cellcolor[HTML]{08773F}} \color[HTML]{F1F1F1} 93.2 & {\cellcolor[HTML]{0A7B41}} \color[HTML]{F1F1F1} 92.0 \\
			\cline{4-9}
			&  &  & $\mathrm{Sel}_{U}$ & {\cellcolor[HTML]{05713C}} \color[HTML]{F1F1F1} 95.5 & {\cellcolor[HTML]{05713C}} \color[HTML]{F1F1F1} 95.7 & {\cellcolor[HTML]{05713C}} \color[HTML]{F1F1F1} 95.4 & {\cellcolor[HTML]{07753E}} \color[HTML]{F1F1F1} 94.1 & {\cellcolor[HTML]{0E8245}} \color[HTML]{F1F1F1} 89.0 \\
			\cline{1-9} \cline{2-9} \cline{3-9} \cline{4-9}
			\multirow[c]{10}{*}{AlpineN2-3D} & \multirow[c]{5}{*}{No GPi} & No Ada &  & {\cellcolor[HTML]{FEFFBE}} \color[HTML]{000000} 0.0 & {\cellcolor[HTML]{FEFFBE}} \color[HTML]{000000} 0.0 & {\cellcolor[HTML]{FEFFBE}} \color[HTML]{000000} 0.0 & {\cellcolor[HTML]{FEFFBE}} \color[HTML]{000000} 0.0 & {\cellcolor[HTML]{FEFFBE}} \color[HTML]{000000} 0.0 \\
			\cline{3-9} \cline{4-9}
			&  & \multirow[c]{2}{*}{Ada} & $\mathrm{Sel}_{\mathrm{Cat}}$ & {\cellcolor[HTML]{1E9A51}} \color[HTML]{F1F1F1} 78.3 & {\cellcolor[HTML]{69BE63}} \color[HTML]{F1F1F1} 59.1 & {\cellcolor[HTML]{78C565}} \color[HTML]{000000} 54.5 & {\cellcolor[HTML]{05713C}} \color[HTML]{F1F1F1} 95.9 & {\cellcolor[HTML]{FEFFBE}} \color[HTML]{000000} 0.0 \\
			\cline{4-9}
			&  &  & $\mathrm{Sel}_{U}$ & {\cellcolor[HTML]{128A49}} \color[HTML]{F1F1F1} 85.5 & {\cellcolor[HTML]{1B9950}} \color[HTML]{F1F1F1} 79.1 & {\cellcolor[HTML]{7DC765}} \color[HTML]{000000} 52.4 & {\cellcolor[HTML]{CFEB85}} \color[HTML]{000000} 23.9 & {\cellcolor[HTML]{FEFFBE}} \color[HTML]{000000} 0.0 \\
			\cline{3-9} \cline{4-9}
			&  & \multirow[c]{2}{*}{iAda} & $\mathrm{Sel}_{\mathrm{Cat}}$ & {\cellcolor[HTML]{B7E075}} \color[HTML]{000000} 33.2 & {\cellcolor[HTML]{E6F59D}} \color[HTML]{000000} 13.2 & {\cellcolor[HTML]{FEE18D}} \color[HTML]{000000} -18.8 & {\cellcolor[HTML]{006837}} \color[HTML]{F1F1F1} 99.6 & {\cellcolor[HTML]{FFFEBE}} \color[HTML]{000000} 0.0 \\
			\cline{4-9}
			&  &  & $\mathrm{Sel}_{U}$ & {\cellcolor[HTML]{42AC5A}} \color[HTML]{F1F1F1} 68.9 & {\cellcolor[HTML]{6BBF64}} \color[HTML]{000000} 58.5 & {\cellcolor[HTML]{98D368}} \color[HTML]{000000} 43.9 & {\cellcolor[HTML]{006837}} \color[HTML]{F1F1F1} 99.7 & {\cellcolor[HTML]{FEFFBE}} \color[HTML]{000000} 0.0 \\
			\cline{2-9} \cline{3-9} \cline{4-9}
			& \multirow[c]{5}{*}{GPi} & No Ada &  & {\cellcolor[HTML]{E2F397}} \color[HTML]{000000} 15.0 & {\cellcolor[HTML]{AFDD70}} \color[HTML]{000000} 36.5 & {\cellcolor[HTML]{FFF0A6}} \color[HTML]{000000} -9.9 & {\cellcolor[HTML]{016A38}} \color[HTML]{F1F1F1} 98.8 & {\cellcolor[HTML]{FFFEBE}} \color[HTML]{000000} 0.0 \\
			\cline{3-9} \cline{4-9}
			&  & \multirow[c]{2}{*}{Ada} & $\mathrm{Sel}_{\mathrm{Cat}}$ & {\cellcolor[HTML]{036E3A}} \color[HTML]{F1F1F1} 97.4 & {\cellcolor[HTML]{04703B}} \color[HTML]{F1F1F1} 96.8 & {\cellcolor[HTML]{04703B}} \color[HTML]{F1F1F1} 96.3 & {\cellcolor[HTML]{006837}} \color[HTML]{F1F1F1} 100.0 & {\cellcolor[HTML]{FEFFBE}} \color[HTML]{000000} 0.0 \\
			\cline{4-9}
			&  &  & $\mathrm{Sel}_{U}$ & {\cellcolor[HTML]{04703B}} \color[HTML]{F1F1F1} 96.3 & {\cellcolor[HTML]{097940}} \color[HTML]{F1F1F1} 92.9 & {\cellcolor[HTML]{06733D}} \color[HTML]{F1F1F1} 94.6 & {\cellcolor[HTML]{016A38}} \color[HTML]{F1F1F1} 99.2 & {\cellcolor[HTML]{FEFFBE}} \color[HTML]{000000} 0.0 \\
			\cline{3-9} \cline{4-9}
			&  & \multirow[c]{2}{*}{iAda} & $\mathrm{Sel}_{\mathrm{Cat}}$ & {\cellcolor[HTML]{0A7B41}} \color[HTML]{F1F1F1} 91.9 & {\cellcolor[HTML]{0B7D42}} \color[HTML]{F1F1F1} 91.1 & {\cellcolor[HTML]{0C7F43}} \color[HTML]{F1F1F1} 90.1 & {\cellcolor[HTML]{006837}} \color[HTML]{F1F1F1} 99.9 & {\cellcolor[HTML]{FFFEBE}} \color[HTML]{000000} 0.0 \\
			\cline{4-9}
			&  &  & $\mathrm{Sel}_{U}$ & {\cellcolor[HTML]{128A49}} \color[HTML]{F1F1F1} 85.8 & {\cellcolor[HTML]{097940}} \color[HTML]{F1F1F1} 92.8 & {\cellcolor[HTML]{51B35E}} \color[HTML]{F1F1F1} 65.4 & {\cellcolor[HTML]{016A38}} \color[HTML]{F1F1F1} 99.2 & {\cellcolor[HTML]{04703B}} \color[HTML]{F1F1F1} 96.8 \\
			\cline{1-9} \cline{2-9} \cline{3-9} \cline{4-9}
			\bottomrule
		\end{tabular}
		\label{tab:wcri-synthetic}
	\end{table}
	
	The collection of reference optimizers considered in Table \ref{tab:wcri-synthetic} is given by
	\begin{equation}
		\Omega=\{\text{BO}(\alpha,\kappa):\alpha\in\{\alpha_{\text{LogEI}},\alpha_{\text{LogPI}},\alpha_{\text{UCB}}\},\kappa\in\{\kappa_{\text{RBF}},\kappa_{\text{Mat.}},\kappa_{\text{RQ}}\}\}
	\end{equation}
	while the BO-GPi, BO-(i)Ada and BO-GPi-(i)Ada optimizer collections that are being compared to $\Omega$ are, respectively:
	\begin{itemize}
		\item $\Omega'=\{\text{BO-GPi}(\alpha):\alpha\in\{\alpha_{\text{LogEI}},\alpha_{\text{LogPI}},\alpha_{\text{UCB}}\}\}$,
		\item $\Omega'=\{\text{BO-(i)Ada}(\kappa,\text{Sel}):\kappa\in\{\kappa_{\text{RBF}},\kappa_{\text{Mat.}},\kappa_{\text{RQ}}\},\text{Sel}\in\{\text{Sel}_{\mathcal{U}},\text{Sel}_{\text{Cat}}\}\}$,
		\item $\Omega'=\{\text{BO-GPi-(i)Ada}(\text{Sel}_{\mathcal{U}}),\text{BO-GPi-(i)Ada}(\text{Sel}_{\text{Cat}})\}$.
	\end{itemize}
	
	From Table \ref{tab:wcri-synthetic}, it can be seen that there is significant confidence that an adaptive BO methodology will at least perform better than the worst-case standard Bayesian optimizer. These results furthermore show that a fully adaptive implementation (BO-GPi-(i)-Ada) leads to better optimization results than the alternatives (BO-(i)Ada, BO-GPi) when faced with the same computational budget.
	
	Lastly, it is important to know about the additional computational expense of adaptive BO over standard BO. This will determine the feasibility of employing it to an engineering problem with an expensive objective. In order to do this, the run times were recorded that give rise to the results in Figure \ref{fig:gpi-iada-performance} and Table \ref{tab:wcri-synthetic}, and the averages were recorded in Table \ref{tab:bo-gpi-iada-runtimes}.
	\begin{table}[H]
		\centering
		\caption{Table of median run times per iteration in seconds. All results were gathered using conventional CPU cores.}
		\label{tab:bo-gpi-iada-runtimes}
		\begin{tabular}{|llll|}
			\hline
			\multirow{2}{*}{Objective} & \multicolumn{2}{l}{BO-GPi-iAda} &
			\multicolumn{1}{l|}{BO}                               \\ \cline{2-4} 
			& \multicolumn{1}{l}{$\text{Sel}_{\text{Cat}}$} & \multicolumn{1}{l}{$\text{Sel}_\mathcal{U}$} & -    \\ \hline
			Sphere-6D                  & \multicolumn{1}{l}{21.20}                     & \multicolumn{1}{l}{21.12}                    & 9.89 \\ \hline
			AlpineN2-3D                & \multicolumn{1}{l}{23.50}                     & \multicolumn{1}{l}{20.33}                    & 7.05 \\ \hline
		\end{tabular}
	\end{table}
	From Table \ref{tab:bo-gpi-iada-runtimes}, it can be seen that the run times of the most extensive adaptive BO scheme, BO-GPi-iAda, are more than double that of standard BO. The relative increase in computational expensive is thus quite substantial. However, expensive FEM simulations could take up one or multiple hours by using the same computational resources. An increase of 10 to 15 seconds per objective evaluation is therefore justified while using adaptive BO.
	
	\section{Problem description and data analysis}
	\label{sec:chip-data}
	
	\subsection{Case Study: Solder Joint Reliability}
	\label{SolverFEM}
	
\begin{figure}[H]
		\centering
		\includegraphics[width=\linewidth]{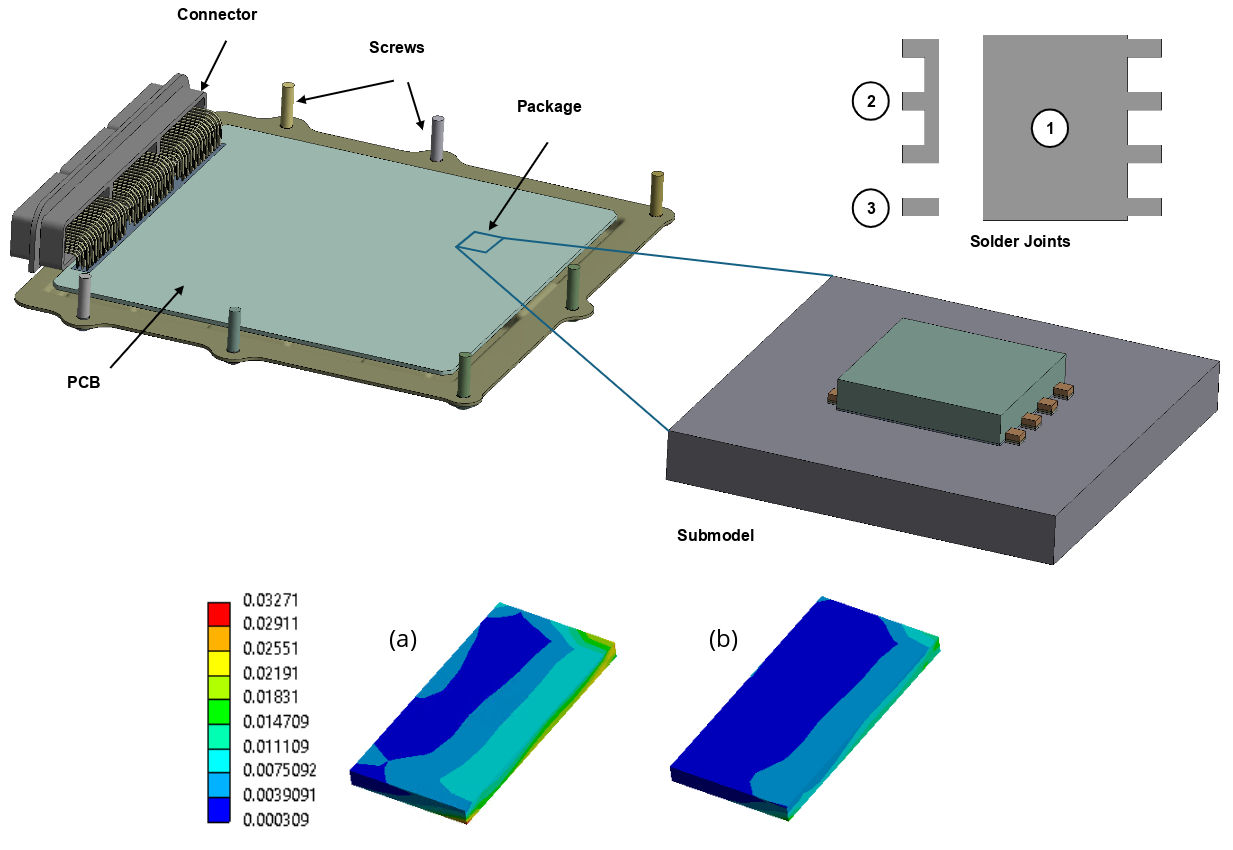}
		\caption{Schematic overview of the PCB model, the electronic package submodel and the solder joint interface (top right) subject to the design problem. Some accumulated creep strain profiles across solder joint 3 are displayed in subfigures (a) and (b).}
		\label{fig:FEM-setup}
	\end{figure}

	In order to put the adaptive BO schemes to the test and to confirm the promising findings that Table \ref{tab:wcri-synthetic} imply, a case study in the automotive power electronics field was considered. The problem statement involves a commercially available printed circuit board (PCB) \cite{niessner2023impact}. The goal is to optimize the materials and the position of the selected package on the available PCB area of approximately \SI{163.4}{\milli\meter} $\times$ \SI{163.4}{\milli\meter} for a minimum accumulated creep strain (dimensionless quantity) in the most critical solder joint of the package under thermal cyclic load. A submodeling-based approach was utilized to solve each case. Figure \ref{fig:FEM-setup} shows the relative footprints of the package and the whole PCB, the submodel of the package-on-PCB assembly, the solder joint layout, and two representative accumulated creep strain profiles.
	
	In the submodel, a \SI{13}{\milli\meter} $\times$ \SI{13}{\milli\meter} area was defined for the PCB around the package, which has a maximum dimension of \SI{6.45}{\milli\meter}. A commercially available software for FEM simulation was utilized to obtain the accumulated creep strain values after multiple temperature cycles between \SI{-40}{} and \SI{125}{\degreeCelsius}. This approach first solves for the displacements of the PCB-only mode, i.e., a global model of the housing and the PCB without any packages mounted on it. The displacement results are then used as the boundary conditions for the submodel --- depending on its location on the PCB --- using the `cut-boundary interpolation' technique along with the same thermal load as the global model. The FEM simulation workflow of the submodel calculates the value of a damage parameter based on the nonlinear accumulated creep strain. This value is volume-based weighted average of the accumulated creep strain over all the finite elements of a solder joint. The result corresponding to the solder joint 3 is selected as the target objective for the optimization problem. This is due to it being the smallest in dimension and, thus, the most critical one. Figure \ref{fig:FEM-setup}(a) and Figure \ref{fig:FEM-setup}(b) show two accumulated creep strain profiles of which the volume-based average needs to be minimized.
	
	Several design parameters were considered for the optimization problem. The geometric parameters include the package position, given by two-dimensional $x$ and $y$ coordinates, and the chip rotation angle, which is fixed at either $0^\circ$ or $90^\circ$. The material parameters include the coefficients of thermal expansion (CTEs) of the molding compound before and after its glass transition. This is indicated by the teal-colored top layer of the package submodel in Figure \ref{fig:FEM-setup}. These are denoted by CTE1 and CTE2 respectively. The selection of material parameters for this study is based on a previous study that shows that among several properties, thermal expansion coefficients of the molding compound affect the stresses in the solder joints the most \cite{gromala2022degradation}. The glass transition temperature ($T_\mathrm{g}$) was defined between \SI{100}{} and \SI{110}{\degreeCelsius}. Considering the serviceable area available on the PCB, the range for the \textit{x} and \textit{y} coordinates of the centroid of the submodel was defined as \SIrange[]{15}{145}{\milli\meter} and \SIrange[]{20}{145}{\milli\meter}, respectively. The rotation was either set to \SI{0}{\degree} or to \SI{90}{\degree}. The range for CTE1 and CTE2 was set to \SIrange[]{5}{12}{ppm/\degreeCelsius} and \SIrange[]{20}{37}{ppm/\degreeCelsius}, respectively. Additional scripting was utilized to automate changing the design parameters, initiating the FEM solver, and extracting the results corresponding to the target objective. A summary of all design parameters can be found in Table \ref{tab:chip-total-design}.
	
	\begin{table}[H]
		\centering
		\caption{Design parameters for the accumulated creep strain optimization problem.}
		\begin{tabular}{|lll|}
			\hline
			Design parameter & Lower bound & Upper bound \\ \hline
			$x$              & \SI{15}{\milli\meter}           & \SI{145}{\milli\meter}          \\ \hline
			$y$              & \SI{20}{\milli\meter}           & \SI{145}{\milli\meter}          \\ \hline
			rotation         & \multicolumn{2}{c|}{\{0°,90°\}} \\ \hline
			CTE1             & 5 ppm/°C    & 12 ppm/°C    \\ \hline
			CTE2             & 20 ppm/°C   & 37 ppm/°C   \\ \hline
		\end{tabular}
		\label{tab:chip-total-design}
	\end{table}
	
	For completeness, the design parameters leading to the accumulated creep strain profiles in Figure \ref{fig:FEM-setup} are given in Table \ref{tab:non-opt}.
			
	\begin{table}[H]
		\centering
		\small
		\caption{Selected design parameters used for Figure \ref{fig:FEM-setup}(a) and \ref{fig:FEM-setup}(b).}
		\begin{tabular}{|llllll|}
			\hline
			Design & CTE1 (ppm/°C) & CTE2 (ppm/°C) & $x$ (mm)   & $y$ (mm)    & Acc. creep (\%) \\ \hline\hline
			(a) & 6    & 27.7 & 58.07 & 108.35 & 0.30      \\ \hline
			(b) & 8.5  & 31.9 & 58.07 & 131.28 & 0.16      \\ \hline
		\end{tabular}
		\label{tab:non-opt}
	\end{table}
	The computational expense of the FEM simulation to yield one single accumulated creep strain value is substantial: between 1.5 and 2 hours. This is orders of magnitude above the reported per-iteration run times with the same computational resources reported in Table \ref{tab:bo-gpi-iada-runtimes} and justifies the application of BO, both the standard and novel schemes.

	\section{Optimization results}
	
	In design problems, it is often reasonable to analyze the design space to avoid potential redundant parameters. Effective tools to this end include design axis projection (``pair-plotting'') and Sobol' sensitivity analysis \cite{nossent2011sobol,zhang2015sobol}. After performing sensitivity analysis, it has been that the 5D problem, as indicated in Table \ref{tab:chip-total-design}, can be reduced to a 3D optimization problem. This takes place by fixing specific values of chip rotation and CTE1, and subsequently optimizing the remaining design parameters. The details can be found in \ref{sec:design-space}. 
	
\begin{figure}[H]
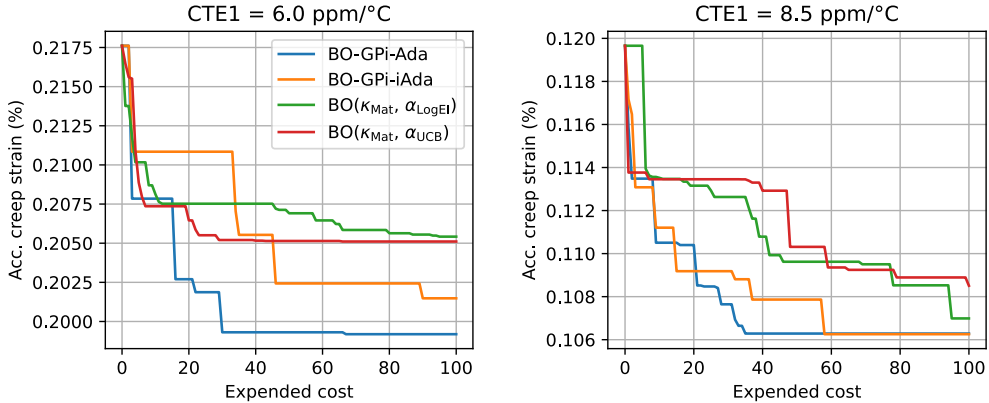

		\centering
		\begin{subfigure}{.45\textwidth}
			\includesvg[width=\textwidth]{img/applications/compas/hist-3D-BO-vs-GPi-Ada-CTE1-6.0ppm.svg}
			\label{fig:compas-3D-opt-6ppm}
		\end{subfigure}
		\begin{subfigure}{.45\textwidth}
			\includesvg[width=\textwidth]{img/applications/compas/hist-3D-BO-vs-GPi-Ada-CTE1-8.5ppm.svg}
			\label{fig:compas-3D-opt-8.5ppm}
		\end{subfigure}
		\caption{Accumulated non-linear creep strain optimization histories, with the incumbent minimized strain value plotted against the expended cost incurred by the optimization scheme indicated by the legend. Each optimization method was allotted with a computational budget of 100 objective evaluations. The adaptive BO schemes were used with a categorical candidate selection strategy at the acquisition step.}
		\label{fig:compas-3D-opt}
	\end{figure}

	In line with the selected values of $0^\circ$ for the rotation value, as well as 6 and 8.5 ppm/°C for CTE1, a variety of BO optimizers have been applied to optimize the accumulated strain value. See Figure \ref{fig:compas-3D-opt}.
	
	The two subfigures of Figure \ref{fig:compas-3D-opt} show that that the kernel and acquisition adaptive BO scheme outperforms both standard BO runs across many of the iterations. It should be noted that the true, global minimum of the accumulated creep strain value across the design space is unknown. Therefore, it cannot be definitively concluded if the synthetic improvement results from Table \ref{tab:wcri-synthetic} are reproduced in the solder joint optimization setting. However, as a direct comparison when CTE1 = 6 ppm/°C, the solder joint design found by BO-GPi-Ada after the optimization budget was expended has a 2.9\% lower accumulated creep than the best-performing vanilla BO scheme. Furthermore, while the improvement of BO-GPi-(i)Ada found compared to BO is only marginal (0.7\%) in the case of CTE1 = 8.5 ppm/°C, the \textit{average} improvement across all iterations is 3.1\%. In other words: adaptive BO will statistically be able to achieve similar levels of improvement with a lower computational budget, confirming the positive outlook that the synthetic results present in Table \ref{tab:wcri-synthetic}.
	
\begin{figure}[H]
		\centering
		\includegraphics[width=.8\textwidth]{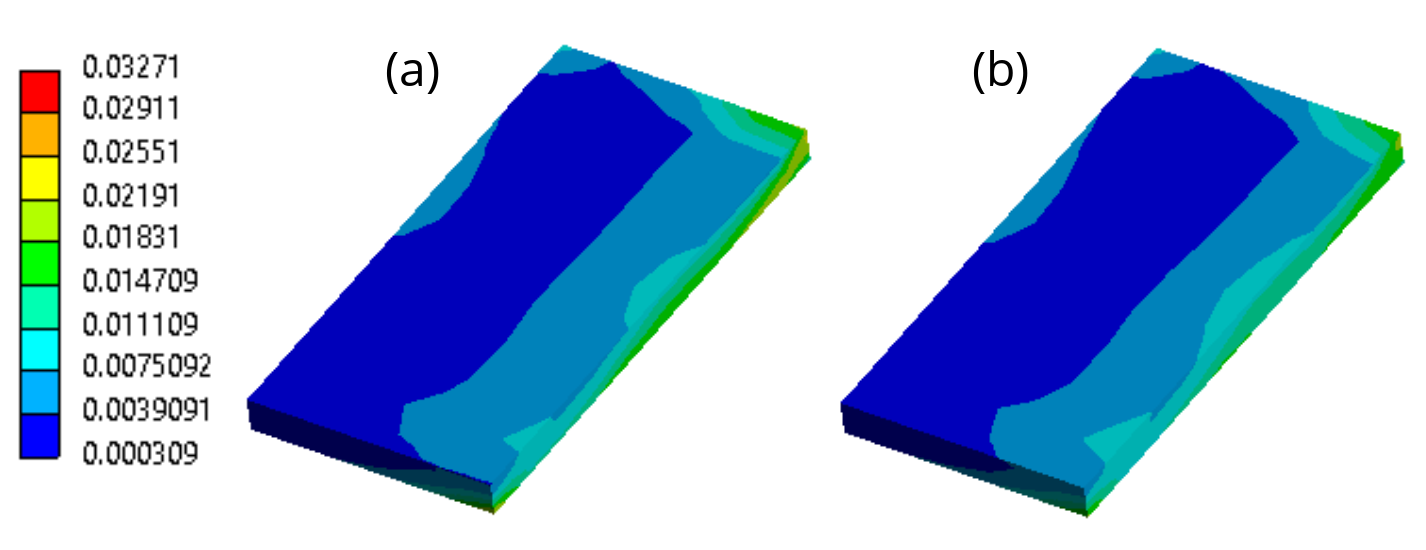}
		\caption{Accumulated creep profiles across the critical solder joint for selected designs (CTE1 = 6 ppm/°C). The profiles correspond to (a) an optimized design by using BO, and (b) an optimized design by using BO-GPi-Ada.}
		\label{fig:compas-3D-opt-designs}
	\end{figure}

\begin{figure}[H]
		\centering
		\includegraphics[width=.8\textwidth]{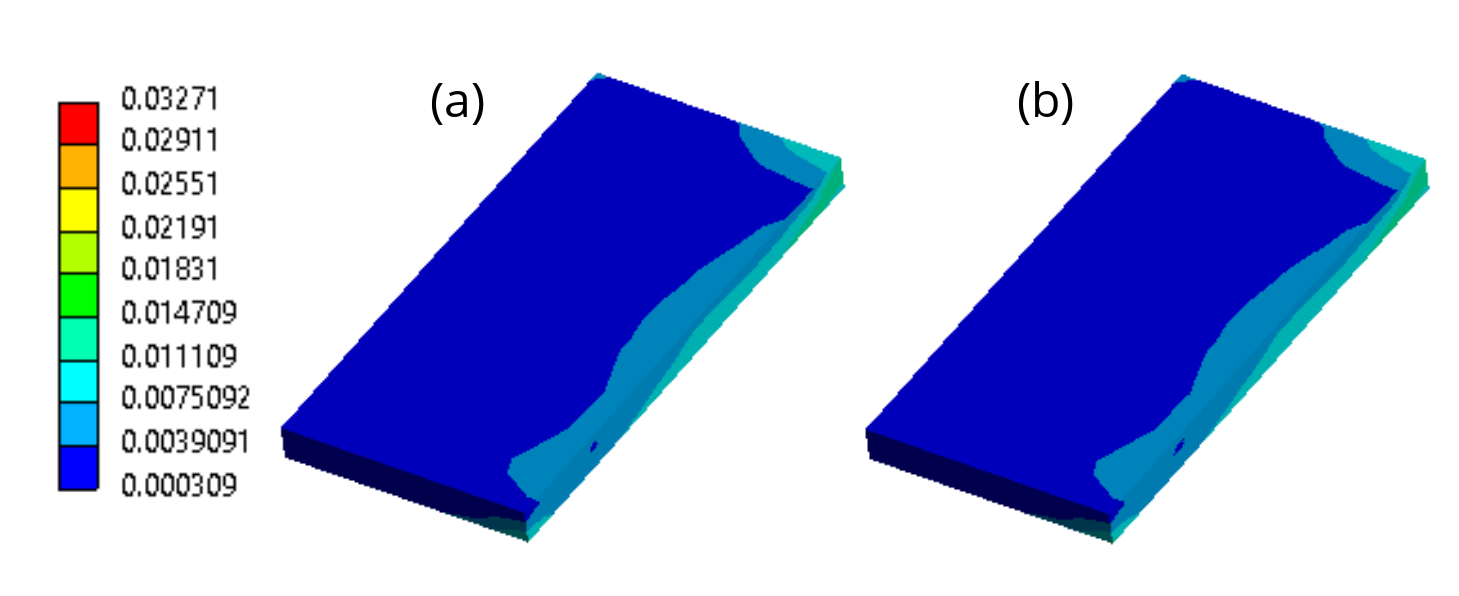}
		\caption{Accumulated creep profiles across the critical solder joint for selected designs (CTE1 = 8.5 ppm/°C). The profiles correspond to (a) an optimized design by using BO, and (b) an optimized design by using BO-GPi-Ada.}
		\label{fig:compas-3D-opt-designs2}
	\end{figure}

	For completeness, the optimized design inputs and objective corresponding to each of the four optimization schemes used in Figure \ref{fig:compas-3D-opt} are recorded in Table \ref{tab:opt-results-6} and Table \ref{tab:opt-results-8.5}. Furthermore, the creep strain profiles of the critical solder joint corresponding to the best performing BO and adaptive BO designs are displayed in Figure \ref{fig:compas-3D-opt-designs} and Figure \ref{fig:compas-3D-opt-designs2}. The design parameters for these strain profiles are given in Table \ref{tab:non-opt}.
	
	\begin{table}[H]
		\centering
		\caption{Optimized design results after 100 iterations (CTE1 = 6 ppm/°C). Best values for standard and adaptive BO schemes are expressed in boldface.}
		\small
		\begin{tabular}{|lllll|}
			\hline
			Opt. method                                              & CTE2 (ppm/°C)         & $x$ (mm)          & $y$ (mm)          & Acc. creep (\%) \\ \hline\hline
			BO($\kappa_{\text{Mat}}$,$\alpha_{\text{UCB}}$) & \textbf{37.0}   & \textbf{69.5} & \textbf{20.0}   & \textbf{0.205}    \\ \hline
			BO($\kappa_{\text{Mat}}$,$\alpha_{\text{LogEI}}$)        & 37.0            & 73.0          & 20.0            & 0.205             \\ \hline
			BO-GPi-Ada                                      & \textbf{36.8} & \textbf{145}  & \textbf{98.9} & \textbf{0.199}    \\ \hline
			BO-GPi-iAda                                              & 35.1          & 145           & 99.2          & 0.202             \\ \hline
		\end{tabular}
		\label{tab:opt-results-6}
	\end{table}
		
	\begin{table}[h]
		\centering
		\caption{Optimized results after 100 iterations (CTE1 = 8.5 ppm/°C). Best values for standard and adaptive BO schemes are expressed in boldface.}
		\small
		\begin{tabular}{|lllll|}
			\hline
			Opt. method                                              & CTE2 (ppm/°C)         & $x$ (mm)          & $y$ (mm)          & Acc. creep (\%) \\ \hline\hline
			BO($\kappa_{\text{Mat}}$,$\alpha_{\text{UCB}}$) & 32.9   & 145 & 10   & 0.109    \\ \hline
			BO($\kappa_{\text{Mat}}$,$\alpha_{\text{LogEI}}$)        & \textbf{35.1}            & \textbf{145}          & \textbf{99.8}            & \textbf{0.107}             \\ \hline
			BO-GPi-Ada                                      & \textbf{37.0} & \textbf{145}  & \textbf{97.8} & \textbf{0.106}    \\ \hline
			BO-GPi-iAda                                              & \textbf{37.0}          & \textbf{145}           & \textbf{98.2}          & \textbf{0.106}             \\ \hline
		\end{tabular}
		\label{tab:opt-results-8.5}
	\end{table}
	
%
	
	 The rows in Table \ref{tab:opt-results-6} and Table \ref{tab:opt-results-8.5} show that the optimizers find (local) minima at various locations in the design space, showing that this design space and objective pose a non-trivial optimization problem. From Figure \ref{fig:compas-3D-opt-designs} and Figure \ref{fig:compas-3D-opt-designs2}, it can be seen that the optimized creep strain profiles are substantially lowered compared to either of the the selected non-optimized profiles in Figure \ref{fig:FEM-setup}(a) and Figure \ref{fig:FEM-setup}(b). Moreover, by comparing Figure \ref{fig:compas-3D-opt-designs}(a) with Figure \ref{fig:compas-3D-opt-designs}(b), it can be seen that the adaptive Bayesian scheme achieves lower accumulated creep strain along the edges of the critical solder joint, confirming the observation from the left subfigure of Figure \ref{fig:compas-3D-opt}. Finally, it should be noted that the profiles found by BO and adaptive BO in the case of CTE1 = 8.5 ppm/°C look very similar, by comparing Figure \ref{fig:compas-3D-opt-designs2}(a) to Figure \ref{fig:compas-3D-opt-designs2}(b). This seems to suggest that the results in using adaptive BO is the same as when using standard BO \cite{wymyslowski2007advanced}. However, as the both subfigures of Figure \ref{fig:compas-3D-opt} show, the optimized result found by BO-GPi-(i)Ada is achieved with at least 50 fewer expensive objective evaluations compared to the (worst-case) BO scheme, which is half of the allotted 100 total evaluation budget. This result shows a significant gain in efficiency when dealing with limited computational resources.
	
	\section{Conclusions}
	Solder joint fatigue due to accumulated creep strain is a possible cause of electronic failure. In this work, simulated behavior of this phenomenon is utilized to do Bayesian data-driven design to minimize the accumulated creep strain. Variance-based Sobol' sensitivity analysis on an identified five-dimensional design space has shown that the dimensionality of this problem is effectively three-dimensional. Optimization results gathered from synthetic objective functions have shown that adaptive BO methods are effective in outperforming (worst-case) standard BO methods with fixed parameters when faced with similar computational budget constraints. These results have been reconfirmed by the application of BO-GPi-(i)Ada to the solder joint design problem. The positive impact of adapting hyperparameters is visible throughout the data-driven design process by achieving an average of 3\% improvement compared to the static hyperparameter alternative. This translates to the ability of reaching similar levels of optimization with much fewer expensive objective evaluations needed.
	
	Future work that could be considered:
	\begin{itemize}
		\item \textbf{Longer run times.} The number of FEM simulations allotted for the BO results has been limited to 100. It is possible for the schemes to find better optima when a larger number of outer loop iterations is considered. This could potentially underline the cost-efficiency of adaptive BO even more clearly.
		\item \textbf{More synthetic results.} The application of the adaptive BO heuristic has delivered promising results. However, the basis on which it was justified, namely the optimization of only two synthetic functions, could be expanded upon. A more extensive set of synthetic objectives should to be considered in order to gain more statistical insights about the performance of adaptive BO. In particular, how do the adaptive schemes perform on classes of objective functions with specific general traits, such as convexity?
		\item \textbf{Different or more complex designs.} The FEM used throughout the design optimization can be expanded in multiple ways. A more complex material model can be used for the molding compound and for calculation of non-linear strain in solder joints, in addition to the accumulated creep strain. Additional geometrical parameters can also be considered, such as the solder standoff height and the dimensions of the molding compound block. It should be remarked that a lot of these considerations will make the design optimization costlier, and thus more attractive to solve with (adaptive) BO.
	\end{itemize}
	
	\section*{Data availability}
	All presented data and the implemented workflow presented in this manuscript are open-source and accessible via GitHub: \url{https://github.com/llguo95/COMPAS_simulation}
	
	\section*{Acknowledgement}
	This work was supported by the ECSEL Joint Undertaking (JU) under Grant 826417. The JU receives support from the European Union’s Horizon 2020 research and innovation program and Germany, Austria, Spain, Finland, Hungary, Slovakia, Netherlands, Switzerland.
	
	A part of this work has been carried out within the project COMPAS, which is supported by ITEA, the Eureka Cluster on software innovation, under project number 19037. COMPAS received funding from Agentschap Innoveren en Ondernemen (Belgium), Bundesministerium für Bildung und Forschung (Germany), and Rijksdienst voor Ondernemend Nederland (Netherlands). 
	
	The authors would like to thank Jiaxiang Yi and dr. Miguel A. Bessa for the valuable discussions and support throughout the implementation and validation of the data-driven methodology. Furthermore, the authors would like to thank Attila Gyarmati and Herbert Güttler from MicroConsult GmbH, Germany for setting up the submodeling-based workflow and dr. Martin Niessner (Infineon) and dr. Michiel van Soestbergen (NXP) for providing the geometrical models of the package and PCB.
	
	\appendix
	\section{Data-driven design space analysis}
	\label{sec:design-space}
	
\begin{figure}[H]
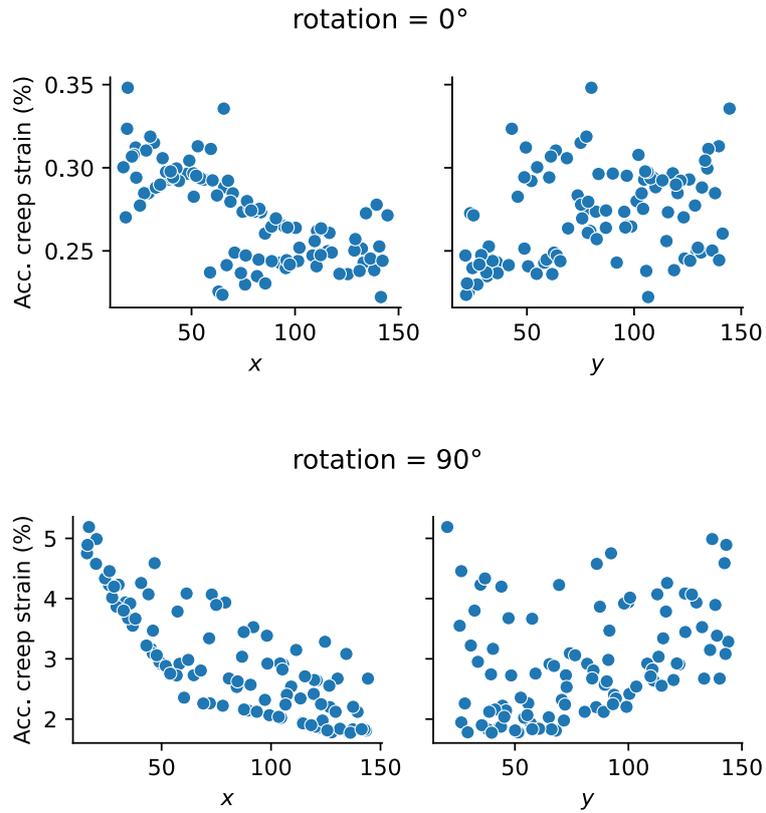

		\centering
		\begin{subfigure}{.7\textwidth}
			\includesvg[width=\textwidth]{img/applications/compas/2D_data_plot_rrotz-0.svg}
			\label{fig:chip-2D-data_plot-rrotz0}
		\end{subfigure}
		\begin{subfigure}{.7\textwidth}
			\includesvg[width=\textwidth]{img/applications/compas/2D_data_plot_rrotz-90.svg}
			\label{fig:chip-2D-data_plot-rrotz90}
		\end{subfigure}
		\caption{The collection of $\pm$100 Saltelli samples (each rotation value) and their accumulated creep strain value projections onto the remaining design parameter axes. For this analysis, fixed values for CTE1 (6 ppm/°C) and CTE2 (30 ppm/°C) were employed.}
		\label{fig:chip-2D-rot-data_plot}
	\end{figure}

	The set of chip rotation values as determined in Table \ref{tab:chip-total-design} is discrete, at either $0^\circ$ or $90^\circ$. Therefore, in order to sample the design space, Saltelli sequence samples are drawn from the remaining design space for each rotation value. A popular data analysis method constitutes plotting the resulting creep strain values against the design parameter axes, which can be seen in Figure \ref{fig:chip-2D-rot-data_plot}.
	
	From Figure \ref{fig:chip-2D-rot-data_plot}, it can be seen that the values of the accumulated creep strain in the $0^\circ$ case are generally an order of magnitude lower when the package is rotated a quarter turn. As the objective is to minimize the strain as much as possible, the original design space can be effectively reduced by fixing the rotation to $0^\circ$.
	
\begin{figure}[H]
		\centering
		\includesvg[width=\textwidth]{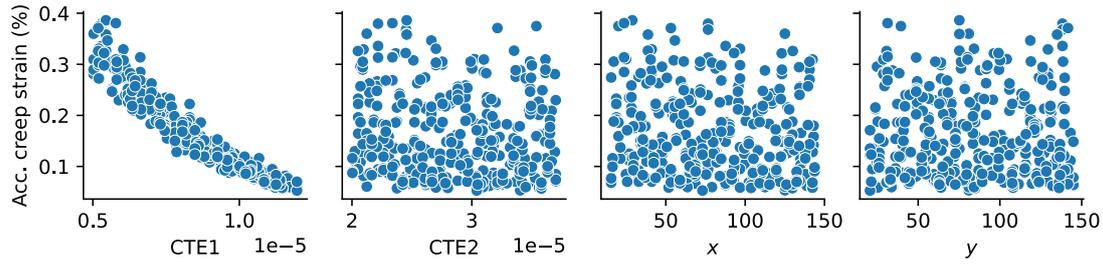}
		\caption{The collection of $354=N_g(D+2)$ Saltelli samples and their accumulated creep strain value projections onto the remaining design parameter axes. Here, $N_g=59$ and $D=4$.}
		\label{fig:chip-4D-data_plot}
	\end{figure}

	For an objective projection plot corresponding to the remaining four design parameters in Table \ref{tab:chip-total-design}, see Figure \ref{fig:chip-4D-data_plot}.
	
\begin{figure}[H]
		\centering
		\begin{subfigure}{.4\textwidth}
			\includesvg[width=\textwidth]{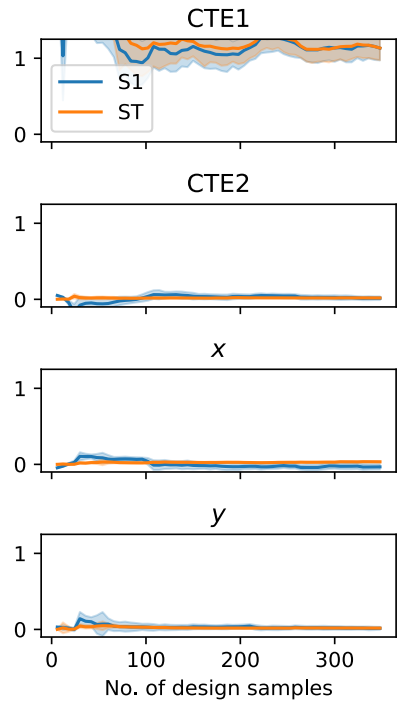}
		\end{subfigure}
		\caption{First- (S1) and total-order (ST) indices and bootstrapped confidence intervals for various numbers of (Saltelli) design samples, corresponding to the design problem with fixed rotation ($0^\circ$).}
		\label{fig:compas-4D-sensitivity-indices-convergence}
	\end{figure}

	From Figure \ref{fig:chip-4D-data_plot}, it can be seen that CTE1 has a clear correlation with the creep strain on the solder joints. It is commonplace for a more thorough design parameter sensitivity analysis to be performed when micro-electronic design is concerned \cite{yang2002parameter,yang2004parametric,vandevelde2003solder}. This design problem is no exception, and the analysis continues by performing Sobol' sensitivity analysis with the obtained Saltelli samples. The sensitivity index convergence plots are shown in Figure \ref{fig:compas-4D-sensitivity-indices-convergence}.
	
	As Figure \ref{fig:compas-4D-sensitivity-indices-convergence} displays, the sensitivity of the strain objective with respect to the remaining parameters is small compared to the CTE1 design parameter. This is an incentive to reduce the 4D problem into a 3D optimization problem by fixing the CTE1 parameter. 
	
\begin{figure}[H]
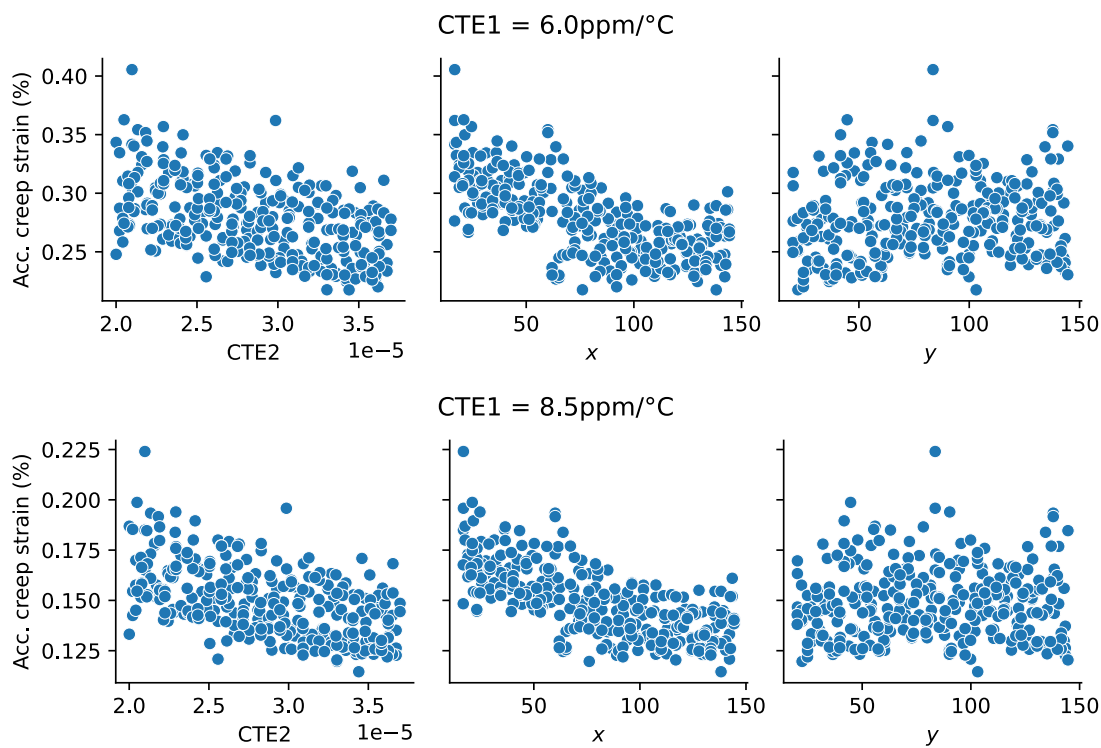

		\centering
		\begin{subfigure}{\textwidth}
			\includesvg[width=\textwidth]{img/applications/compas/3D_CTE1_6ppm_data_plot.svg}
		\end{subfigure}
		\begin{subfigure}{\textwidth}
			\includesvg[width=\textwidth]{img/applications/compas/3D_CTE1_8.5ppm_data_plot.svg}
		\end{subfigure}
		\caption{Each row shows a collection of $310=N_g(D+2)$ Saltelli samples and their accumulated creep strain value projections onto the remaining design parameter axes for different fixed values of CTE1. Here, $N_g=62$ and $D=3$.}
		\label{fig:chip-3D-data_plot}
	\end{figure}

	In order to analyze the residual problem, a similar data analysis is done on the basis of a number of nominal values for CTE1. In order to select these values, note that there is a larger perceived variance of the data when the value of CTE1 is on the lower end of the defined range (leftmost subfigure of Figure \ref{fig:chip-4D-data_plot}). Hence, two nominal values for CTE1, being 6 ppm/°C and 8.5 ppm/°C, are decided upon. See Figure \ref{fig:chip-3D-data_plot} for the accumulated creep strain samples projected on the remaining design parameter axes.
	
	From Figure \ref{fig:chip-3D-data_plot}, it should be noticed that the behavior of the creep strain as a function of the design parameters CTE2, $x$ and $y$ are very similar across the various CTE1 values. However, this behavior is exhibited on a different output scale, with higher CTE1 values corresponding to a lower creep strain. This is intuitively clear from the first subfigure in Figure \ref{fig:chip-4D-data_plot}.
	
\begin{figure}[H]
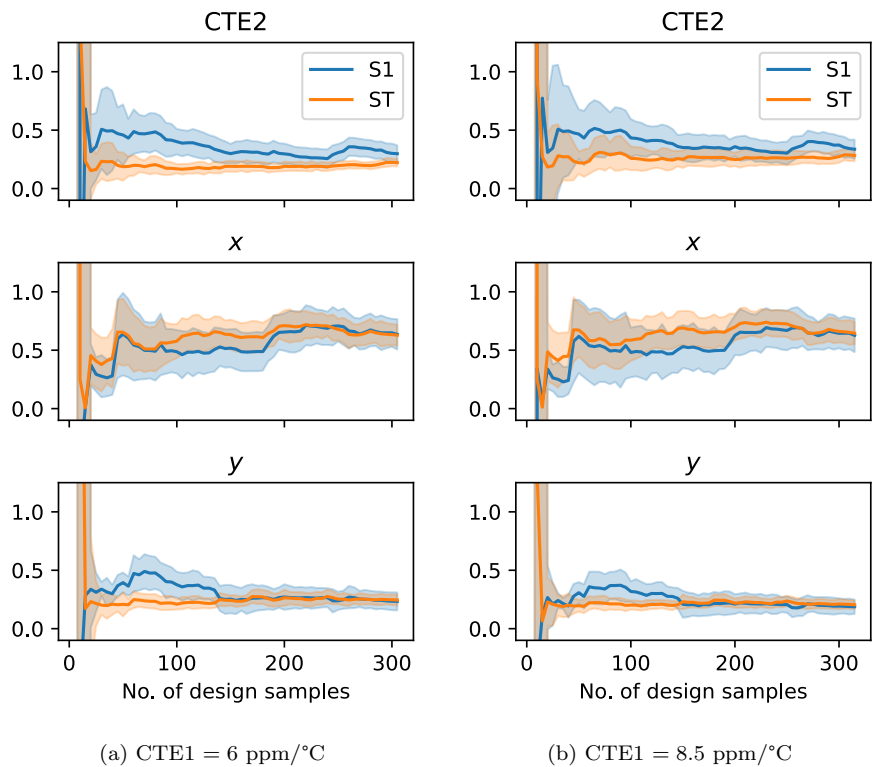

		\centering
		\begin{subfigure}{.4\textwidth}
			\includesvg[width=\textwidth]{img/applications/compas/3D_CTE1_6ppm-ssa-conv.svg}
			\caption{CTE1 $=6$ ppm/°C}
		\end{subfigure}
		\begin{subfigure}{.4\textwidth}
			\includesvg[width=\textwidth]{img/applications/compas/3D_CTE1_8.5ppm-ssa-conv.svg}
			\caption{CTE1 $=8.5$ ppm/°C}
		\end{subfigure}
		\caption{First- (S1) and total-order (ST) indices and bootstrapped confidence intervals for various numbers of (Saltelli) design samples, corresponding to the design problem with fixed rotation ($0^\circ$) and fixed CTE1.}
		\label{fig:compas-3D-sensitivity-indices-convergence}
	\end{figure}

	The fact that the sensitivity profile is largely similar on the reduced design space is of importance, because this justifies the representation of the entire CTE1 domain by virtue of fixing one or a few CTE1 values. To this end, one should confirm the heuristic that altering CTE1 does not influence the sensitivity of the objective with respect to the remaining parameters. To confirm this, the Saltelli samples have been used to construct a Sobol' sensitivity index convergence plot on the residual design space in Figure \ref{fig:compas-3D-sensitivity-indices-convergence}.
	
	Figure \ref{fig:compas-3D-sensitivity-indices-convergence} reveal that there is no significant alteration of the variance-based sensitivity profile when higher values of CTE1 are used. As mentioned previously, this fact allows for the dimensional reduction of the design problem by keeping CTE1 at fixed values when performing the optimization routine.
	
	\bibliographystyle{elsarticle-num}
	\bibliography{dissertation.bib}
	
\end{document}